\pdfoutput=1

\documentclass[11pt]{article}

\usepackage[preprint]{acl}

\usepackage{times}
\usepackage{latexsym}
\usepackage{amssymb}

\usepackage[T1]{fontenc}

\usepackage[utf8]{inputenc}

\usepackage{microtype}

\usepackage{inconsolata}

\usepackage{graphicx}

\usepackage{booktabs}
\usepackage{multirow}
\usepackage{caption}
\usepackage{makecell}
\usepackage{booktabs}
\usepackage{xcolor}
\usepackage[breakable]{tcolorbox}
\usepackage{enumitem}
\usepackage{float}

\definecolor{darkgreen}{rgb}{0.0, 0.5, 0.0}

\NewDocumentCommand{\heng}
{ mO{} }{\textcolor{red}{\textsuperscript{\textit{Heng}}\textsf{\textbf{\small[#1]}}}}

\NewDocumentCommand{\xiusi}
{ mO{} }{\textcolor{red}{\textsuperscript{\textit{Xiusi}}\textsf{\textbf{\small[#1]}}}}

\title{Beyond Sample-Level Feedback: Using Reference-Level Feedback to Guide Data Synthesis}

\author{
Shuhaib Mehri \quad Xiusi Chen \quad Heng Ji \quad Dilek Hakkani-T\"ur  \\ \normalfont{University of Illinois Urbana-Champaign} \\ \texttt{\{mehri2, xiusic, hengji, dilek\}@illinois.edu}
}

\begin{document}
\maketitle

\begin{abstract}  
High-quality instruction-tuning data is crucial for developing Large Language Models (LLMs) that can effectively navigate real-world tasks and follow human instructions. While synthetic data generation offers a scalable approach for creating such datasets, it imposes a quality ceiling where models trained on the data cannot outperform the LLM generating it. To overcome this limitation, we introduce \textsc{Reference-Level Feedback}, a paradigm that extracts desirable characteristics from carefully curated reference samples to guide the synthesis of higher-quality instruction-response pairs. Using this approach, we synthesize \textsc{REFED}, a dataset of 10K instruction-response pairs. Fine-tuning Llama-3.1-8B-Instruct and Mistral-7B-Instruct on \textsc{REFED} demonstrate state-of-the-art performance among similarly sized models, notably reaching a 43.96\% length-controlled win-rate on AlpacaEval 2.0. Extensive experiments demonstrate that \textsc{Reference-Level Feedback} consistently outperforms traditional sample-level feedback methods, generalizes across model architectures, and produces high-quality and diverse data at low cost.
\end{abstract}
\begin{figure}[t]
    \centering
    \includegraphics[width=\linewidth, trim=0 0 0 0pt, clip]{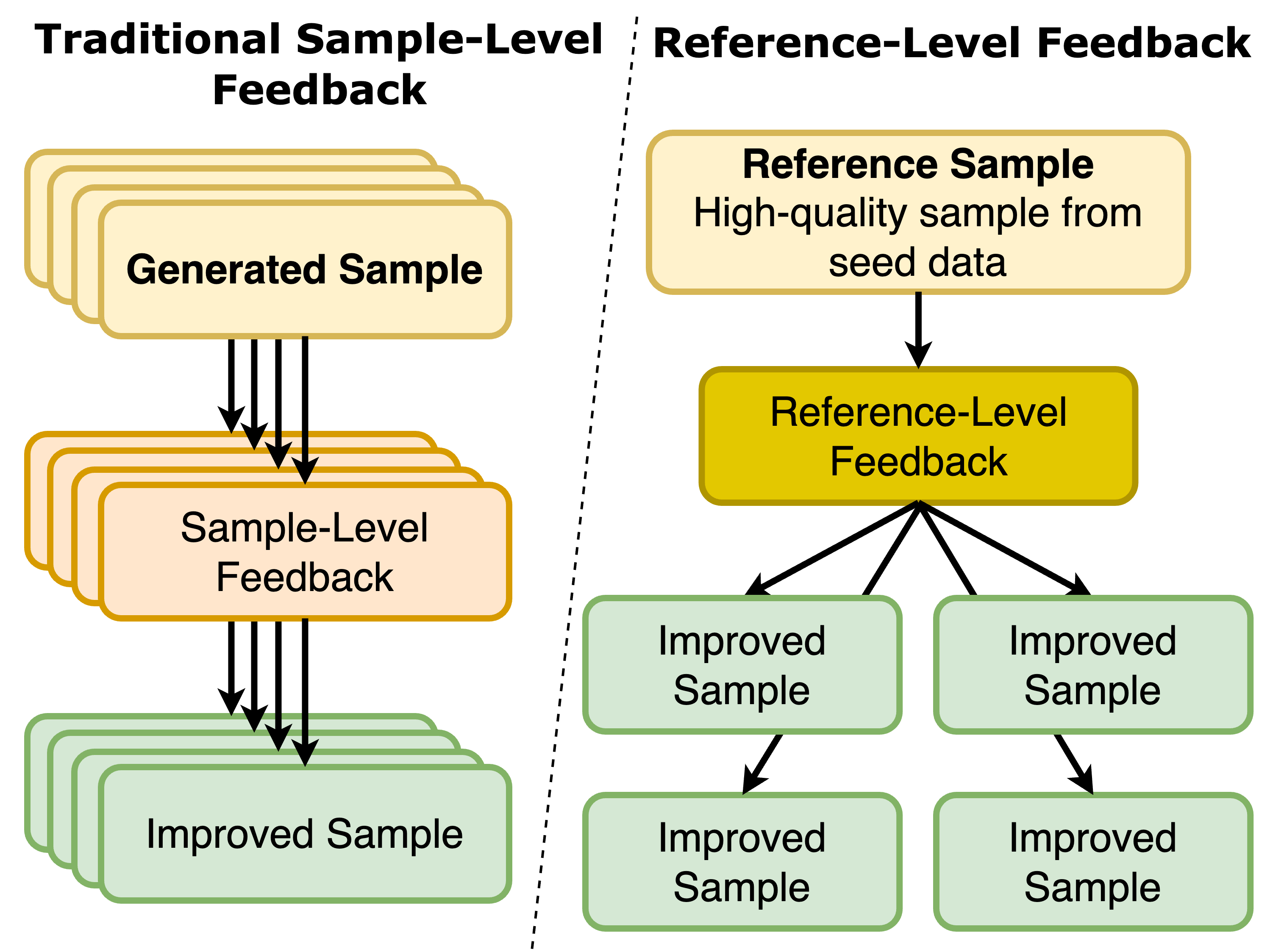} \hfill

    \vspace{-15pt}   
    \caption{Comparison of feedback approaches for data synthesis. Left: Traditional sample-level feedback generates and applies feedback individually for each sample. Right: Our \textsc{Reference-Level Feedback} approach collects feedback once from a high-quality reference sample and applies it multiple new samples.
    }
    \label{fig:intro}

    \vspace{-15pt}
\end{figure}

\section{Introduction}

Large Language Models (LLMs) that can effectively navigate real-world tasks and follow human instructions must be trained on a foundation of high-quality instruction-tuning data that reflect the patterns and characteristics we aim to propagate \citep{NEURIPS2022_b1efde53,bai2022constitutionalaiharmlessnessai}. Though human annotation can produce such high-quality data \citep{NEURIPS2022_b1efde53,lima}, it is not scalable due to its expensive and inefficient nature, making synthetic data generation the practical alternative for large-scale dataset creation. At its core, the quality of synthetic data is bounded by the capabilities of the LLM generating it, imposing a quality ceiling where models trained on this data cannot surpass the data generator's performance \citep{gudibande2023false}.

To improve data quality, recent approaches have incorporated feedback mechanisms that refine generated samples through self-reflection or critique from stronger models \citep{chen2024iteralign,chen2024learning, sun2024principle, bai2022constitutionalaiharmlessnessai}. These sample-level feedback methods have proven effective in improving LLM performance on alignment benchmarks and reinforcing principles like helpfulness and truthfulness \citep{chen2024learning, sun2024principle, bai2022constitutionalaiharmlessnessai}, however they still operate within the quality ceiling of the generating LLM.

Meanwhile, existing synthetic data generation approaches carefully curate seed datasets of high-quality instruction-response pairs, but only use them as in-context examples, underutilizing the rich quality signals they contain \citep{wang-etal-2023-self-instruct, xu2023wizardlmempoweringlargelanguage, alpaca}. To fully leverage these quality signals, we introduce \textsc{Reference-Level Feedback} (see Figure \ref{fig:intro}), a paradigm that explicitly extracts desirable characteristics from high-quality reference samples and systematically propagates them throughout the synthesis process.

Our framework is presented in Figure \ref{fig:pipeline}. For each reference sample, we identify the desirable characteristics of both the instruction and response components, and use them to create feedback for each component. The instruction-specific feedback is used to guide the synthesis of new instructions, which we generate corresponding responses for. Then, the response-specific feedback is used to refine the responses. Since synthesized instructions share key characteristics of their reference counterparts, response-specific feedback remains relevant and is used to improve the quality of synthesized responses. By systematically propagating the desirable characteristics of reference samples to newly generated data, our framework enables us to break through the quality ceiling that normally bounds synthetic data to the generator's capabilities.

We demonstrate the effectiveness of our approach through \textbf{RE}ference-Level \textbf{F}eedback \textbf{E}nhanced \textbf{D}ata (\textsc{REFED}), a 10K-sample dataset synthesized using our framework, with LIMA \citep{lima} as the seed dataset. Llama-3.1-8B-Instruct and Mistral-7B-Instruct fine-tuned on \textsc{REFED} demonstrate significant improvements on AlpacaEval 2.0 \citep{dubois2024lengthcontrolledalpacaevalsimpleway} and Arena-Hard \citep{arenahard2024}. Notably, we achieve state-of-the-art performance among similarly sized models on AlpacaEval 2.0, with Llama-3.1-8B-Instruct reaching 43.96\% length-controlled win rate. Through comprehensive experiments, we demonstrate that \textsc{Reference-Level Feedback} (1) outperforms traditional sample-level feedback approaches, (2) generalizes across different model architectures and improves both base and instruct variants, (3) produces synthetic data higher diversity scores than existing datasets, and (4) achieves these results cost-effectively, synthesizing 10K high-quality samples for under \$20 using GPT-4o mini.

The main contributions of this work are \footnotemark[1]:
\begin{itemize}[leftmargin=10pt,topsep=0pt]
\itemsep -0.5ex
    \item We introduce \textsc{Reference-Level Feedback} for data synthesis, a method that captures desirable characteristics in reference samples and propagates it to newly synthesized data through feedback. Using our framework, and the LIMA \citep{lima} training dataset as seed data, we synthesize \textsc{REFED}, a dataset of 10K instruction-response pairs.
    \item We demonstrate the effectiveness of our approach by achieving strong performance on AlpacaEval 2.0 and Arena-Hard \citep{dubois2024lengthcontrolledalpacaevalsimpleway,arenahard2024}. Notably, we achieve state-of-the-art performance on the AlpacaEval 2.0 leaderboard, with a 21.06\% improvement in length-controlled win rate.
    \item We conduct comprehensive experiments that demonstrate that our approach outperforms models trained on other synthetic instruction-tuning datasets, consistently improves base and instruct variants of different model architectures, and provides more effective quality improvements compared to traditional sample-level feedback approaches, while also being more efficient.
\end{itemize}

\footnotetext[1]{Our code and data are available at \url{https://anonymous.4open.science/r/anon_refed-DD20}}

\section{Related Work}

\paragraph{Synthetic Data for Instruction Tuning.}

Data synthesis has emerged as an effective and scalable approach to creating instruction-tuning datasets. One line of approaches use instruction-response pairs from a seed dataset as in-context examples to guide synthesis \citep{wang-etal-2023-self-instruct, alpaca, peng2023instructiontuninggpt4}. While \citet{wang-etal-2023-self-instruct} use models to self-generate their training data, subsequent works leverage more capable proprietary models to generate higher quality data \citep{alpaca, peng2023instructiontuninggpt4}.

Other approaches have explored alternative synthesis strategies. There are works that use structured guidance through manually curated taxonomies or LLM-generated skill sets \citep{li2024syntheticdataalmostscratch, kaur2024instruct}. \citet{xu2024magpiealignmentdatasynthesis} uses pre-query templates to sample instructions from aligned LLMs and generate instructions that reflect the LLM's existing knowledge.

Many works have explored methods for enhancing the quality of synthesized data. \citet{xu2023wizardlmempoweringlargelanguage} proposes Evol-Instruct, which generates increasingly complex versions of existing instructions. Other approaches include using multi-agent simulation \citep{tang2024synthesizingposttrainingdatallms} or incorporating natural language feedback \citep{bai2022constitutionalaiharmlessnessai, chen2024learning, NEURIPS2023_0764db11}.

\paragraph{Natural Language Feedback.}

Using LLMs to evaluate natural language text has become a standard practice in recent years \citep{mehri2023automatic}. Beyond evaluating, natural language feedback is a rich medium for conveying detailed and nuanced information to guide model improvement, and recent studies have demonstrated its effectiveness in improving LLM performance. \citet{NEURIPS2023_91edff07} introduce Self-Refine, which has LLMs generate feedback and refine their own responses. Following this, several works have shown that using various feedback methods and fine-tuned critic models can yield further improvements \citep{jin2023dataefficient, wang2023shepherdcriticlanguagemodel, gou2024criticlargelanguagemodels, wu2024meta}.

Another application of feedback is at the dataset level, focusing on creating higher-quality training data. Constitutional AI \citep{bai2022constitutionalaiharmlessnessai} generates self-critiques and revisions to create training data aligned with specific principles. In a similar manner, Self-Align \citep{NEURIPS2023_0764db11} uses natural language descriptions of desirable qualities to guide LLMs towards producing more aligned outputs, IterAlign \citep{chen2024iteralign} uses an iterative process to discover constitutions and self-correct, and \citet{chen2024learning} demonstrate the effectiveness of feedback-based refinement in code generation.

In order to more effectively incorporate feedback for data synthesis, we introduce \textsc{Reference-Level Feedback}. It fundamentally differs from existing feedback-based methods in three key aspects. First, while previous work collects feedback at the sample-level, we collect feedback from high-quality reference samples in the seed data. This enables us to identify and propagate desirable qualities from reference samples and establish higher quality standards for the data synthesis.

Secondly, our approach effectively leverages seed datasets. Rather than using seed data samples as in-context examples for synthesizing similar samples, we systematically analyze and explicitly capture its desirable qualities

Lastly, we expand the role of feedback beyond response refinement and guide the entire data synthesis process: our method uses feedback to synthesize new instructions in addition to refining the corresponding responses.
\section{Method}

\begin{figure*}[t]
\vspace{-15pt}
    \centering
    \includegraphics[width=\linewidth, trim=0 100 0 0pt, clip]{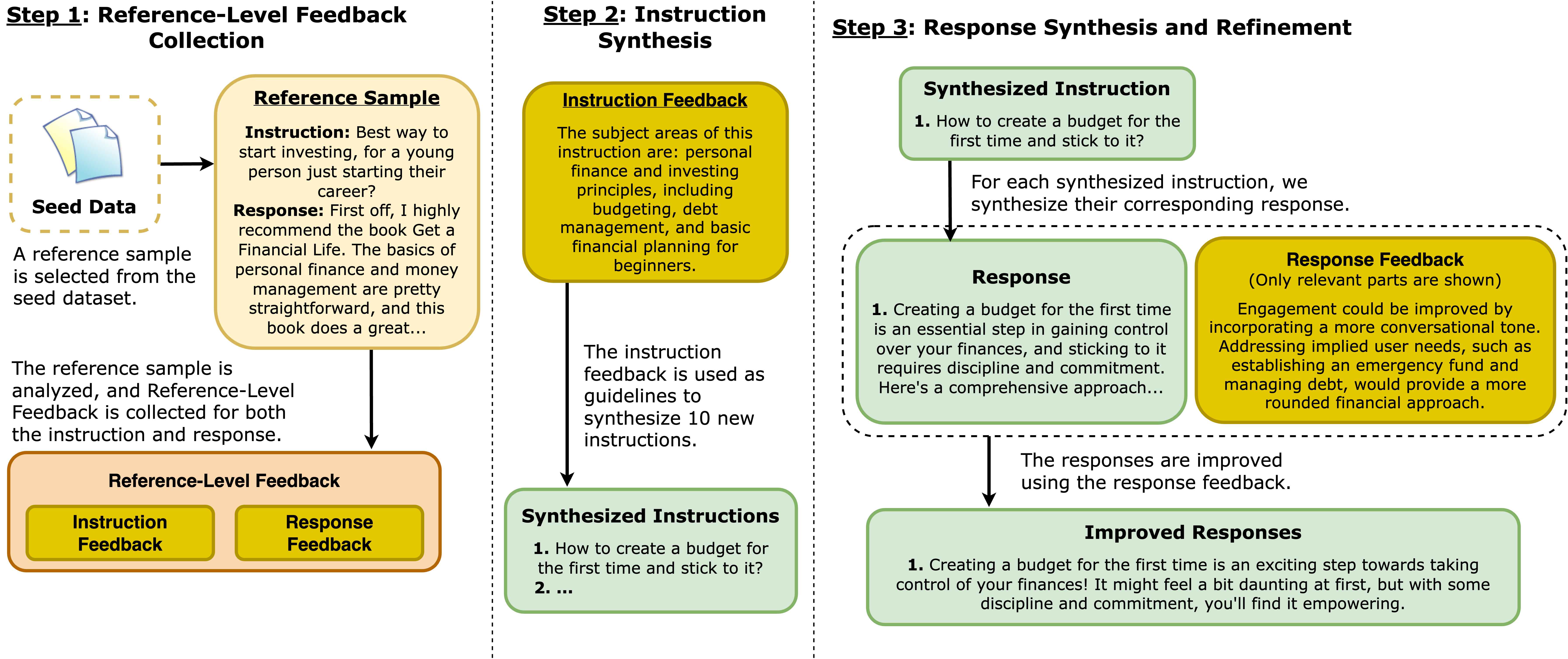} \hfill

    \vspace{-15pt}
    
    \caption{An overview of our data synthesis pipeline. Starting from our seed data, we select a reference sample and collect \textsc{Reference-Level Feedback} on both the instruction and response. Instruction feedback is used to synthesize new instructions. We generate corresponding responses, and then improve it using response feedback.}
    \label{fig:pipeline}

\vspace{-15pt}
\end{figure*}

In this section, we present our data synthesis pipeline that leverages \textsc{Reference-Level Feedback} to generate high-quality instruction-response pairs. An overview of the pipeline is presented in Figure \ref{fig:pipeline}, and the steps are detailed in the following subsections. Complete examples for each step can be found in Appendix \ref{sec:appendix_examples}, and the prompts used for each section can be found in Appendix \ref{sec:appendix_prompt_templates}.

\subsection{Feedback Collection}

Our pipeline begins with a seed dataset -- a small collection of carefully curated instruction-response pairs that serve as exemplars for synthesized data samples. It can be either manually crafted by human annotators or automatically selected using quality-based criteria. These reference samples are high-quality and exhibit desirable characteristics such as clarity and relevance, which we aim to replicate in our synthetic data. For \textsc{Reference-Level Feedback}, we systematically identify and capture such qualities through a framework that identifies the strength of each sample, as well as potential areas for improvement.

Unlike traditional approaches that collect feedback on generated responses at the sample-level, our method identifies the qualities that make reference samples high-quality and uses it for feedback. This feedback captures a richer signal than feedback collected at the sample-level, establishing higher quality standards for synthesis and providing more effective guidance for generating training data that exhibits similar properties to the reference samples.

For each reference sample in the seed dataset, we collect \textsc{Reference-Level Feedback} from both the instruction and the response:

\textbf{Instruction Feedback.} We analyze key attributes (e.g. clarity and actionability) of reference instructions and capture essential features that make it effective for training. We also ensure comprehensive coverage along a wide breadth by collecting feedback along two dimensions: relevant subject areas (e.g. cellular biology, csv file manipulation, legislative processes) and relevant skills necessary to respond to the instruction (e.g. understanding of specific tools, knowledge of processes, analysis). This enables us to systematically identify desirable characteristics of instructions while maximizing the breadth of instruction types.

\textbf{Response Feedback.} When collecting feedback from a reference response, we identify key qualities that make it an effective response to the instruction. We evaluate along multiple critical dimensions, including factual accuracy, relevance to the instruction, and comprehensiveness. This feedback captures both the strengths of the reference response and specific areas where it can be improved upon.

\subsection{Data Synthesis}
Now, we use the collected \textsc{Reference-Level Feedback} from the previous stage to synthesize new data samples, while maintaining the quality standards established by our reference data. For each reference sample and its corresponding feedback, we employ a two-phase synthesis process, as illustrated in Figure \ref{fig:pipeline}:

\begin{enumerate}
    \item \textbf{Instruction Synthesis.} We provide an LLM the reference instruction as an example and the instruction feedback as guidelines to synthesize new instructions that maintain the qualities specified in the feedback. As depicted in Step 2 of Figure \ref{fig:pipeline}, we synthesize 10 new instructions for \textbf{subject-based} feedback, which produces instructions that align with the subject areas of the reference response. We also synthesize 10 new instructions for \textbf{skill-based} feedback, which produces instructions that align with the skills needed to respond to the reference instruction.
    
    \item \textbf{Response Synthesis and Refinement.} For each synthesized instruction, we first generate an initial response. We then enhance this response using the reference response feedback, instructing the language model to analyze the feedback and incorporate the relevant aspects. This process is shown in Step 3 of Figure \ref{fig:pipeline}.
    
    \paragraph{Note on relevance of response feedback.}
    Although the response feedback was originally collected for the reference response, many aspects of it can still remain applicable because of the shared characteristics between the reference and synthesized instructions. We acknowledge that not all feedback elements may transfer, and to account for this, we explicitly instruct the model to selectively apply only the relevant aspects of the feedback and ignore the irrelevant aspects. An example of this can be found in \ref{sec:appendix_examples}.
\end{enumerate}

This synthesis process enables us to synthesize new data, while systematically propagating the high-quality characteristics of reference samples.

\begin{table*}[t]
    \centering
    \small
    \setlength{\tabcolsep}{4pt}
    
    \begin{tabular}{c l c c c c c c} \toprule
        \multicolumn{2}{c}{\multirow{2}[3]{*}{\makecell{\textbf{Feedback Type}}}}
        & \multicolumn{4}{c}{\textbf{AlpacaEval 2}} & \multicolumn{2}{c}{\textbf{Arena-Hard}} \\ \cmidrule(lr){3-6} \cmidrule(lr){7-8}
        & & LC (\%) & WR (\%) & SE & Len. & WR (\%) & Tok. \\ \midrule
        
        & No Feedback & $32.45_{\textcolor{darkgreen}{\uparrow 9.55}}$ & $32.98_{\textcolor{darkgreen}{\uparrow 9.54}}$ & 1.65 & 2106 & $29.2_{\textcolor{darkgreen}{\uparrow 7.9}}$ & 873 \\ 
        & \textsc{Ref-Level} Instruction Feedback & $38.99_{\textcolor{darkgreen}{\uparrow 16.09}}$ & $35.34_{\textcolor{darkgreen}{\uparrow 11.90}}$ & 1.68 & 1926 & $29.8_{\textcolor{darkgreen}{\uparrow 8.5}}$ & 634 \\
        & \textsc{Ref-Level} Instruction + Sample-Level Response Feedback & $42.92_{\textcolor{darkgreen}{\uparrow 20.02}}$ & $41.74_{\textcolor{darkgreen}{\uparrow 18.30}}$ & 1.73 & 1959 & $30.8_{\textcolor{darkgreen}{\uparrow 9.5}}$ & 642 \\ 
        & \textbf{\textsc{Ref-Level} Instruction + Response Feedback} & \textbf{43.96$_{\textcolor{darkgreen}{\uparrow 21.06}}$} & \textbf{42.24$_{\textcolor{darkgreen}{\uparrow 18.80}}$} & \textbf{1.74} & \textbf{1950} & \textbf{35.9$_{\textcolor{darkgreen}{\uparrow 14.6}}$} & \textbf{670} \\ 
        
        \bottomrule
    \end{tabular}
    
    \caption{ Analysis of the different components of \textsc{Reference-Level Feedback} for data synthesis. We evaluate the impact of the instruction and response feedback, and also compare against traditional sample-level feedback for response improvement, while using reference samples from LIMA \citep{lima}. Results show performance after finetuning Llama-3.1-8B-Instruct on each generated dataset. Green subscripts indicate improvements after fine-tuning. Metrics shown are: Length-Controlled Win Rate (LC), Win Rate (WR), Standard Error (SE), Average Length (Len.), and Average \# Tokens (Tok.).}

    \label{tab:table1}
\end{table*}

\begin{table*}[t]
    \centering
    \small
    \setlength{\tabcolsep}{4pt}
    
    \begin{tabular}{c l c c c c c c c} \toprule
        \multicolumn{2}{c}{\multirow{2}[3]{*}{\makecell{\textbf{Model} }}}
        & {\multirow{2}[3]{*}{{\textbf{\# Samples}}}}
        & \multicolumn{4}{c}{\textbf{AlpacaEval 2}}
        & \multicolumn{2}{c}{\textbf{Arena-Hard}} 
        \\ 
        \cmidrule(lr){4-7} \cmidrule(lr){8-9}
        & & & LC (\%) & WR (\%) & SE & Len. & WR (\%) & Tok. \\ \midrule        
        & Llama-3.1-8B-Instruct & - & 22.90 & 23.44 & 1.49 & 2181 & 21.3 & 861 \\ 
        & \quad + Alpaca & 52K & 10.80 & 4.60  & 0.72 & 530 & 6.6 &  321 \\ 
        & \quad +  Evol Instruct & 143K & 13.65 & 6.77  & 0.88 & 949 &  7.0 & 532 \\ 
        & \quad +  UltraChat & 208K & 13.57 & 6.52 &  0.86  & 853 & 7.8 & 500 \\
        & \quad +  Instruct-SkillMix  & 4K & 43.31 & 38.43  & 1.71 &  1658  &  25.2 &  466 \\ 
         \midrule
        & Infinity-Instruct-7M-Gen-Llama3.1-8B & 9M & 31.62 & 25.78  &  1.54 &  1588  & 33.1 & 716 \\ 
        & Llama-3-8B-Instruct-SkillMix & 4K & 38.63 & \textbf{42.98} & 1.75 &  4058  & 12.8 & 1790 \\ 

        & Gpt-3.5-turbo-0613 & - & 22.35 & 14.10 & 1.04 & 1331 & 24.8 & 401 \\
        & Llama-3.1-405B-Instruct & - & 39.26 & 39.11 & 1.43 & 1988 & \textbf{69.3} & 658 \\
        & Claude 3 Opus (02/29) & - & 40.51 & 29.11 & 1.39 & 1388 & 60.4 & 541 \\
        
        \midrule
        & \textbf{Llama-3.1-8B-Instruct-\textsc{REFED}} & 10K & \textbf{43.96}  &  42.24  &  1.74  &  1950  &  35.9 & 670\\
        \bottomrule
    \end{tabular}
    
    \caption{ Evaluation results of Llama-3.1-8B-Instruct finetuned on \textsc{REFED} against selected baselines (detailed in Section \ref{sec:4.2.2}). \textit{Top} shows results from finetuning on various synthetic datasets. \textit{Middle} shows the performance of leading models from AlpacaEval 2.0 leaderboard. \textit{Bottom} shows our model trained on \textsc{REFED}. Results demonstrate that our model outperforms these baselines across both evaluation benchmarks.
    }
    \label{tab:table2}
    \vspace{-15pt}
\end{table*}
\begin{table*}[t]
    \centering
    \small
    \setlength{\tabcolsep}{4pt}
    
    \begin{tabular}{c l c c c c c c} \toprule
        \multicolumn{2}{c}{\multirow{2}[3]{*}{\makecell{\textbf{Model} \\ }}}
        & \multicolumn{4}{c}{\textbf{AlpacaEval 2}} & \multicolumn{2}{c}{\textbf{Arena-Hard}} \\ \cmidrule(lr){3-6} \cmidrule(lr){7-8}
        & & LC (\%) & WR (\%) & SE & Len. & WR (\%) & Tok. \\ \midrule

        & Mistral-7B-v0.3 & - & - & - & - & - & - \\ 
        & \quad + \textsc{REFED} & 16.97 & 17.70 &  1.34  &  2070  & 3.6 & 669 \\ 

        \midrule
        & Mistral-7B-Instruct-v0.3 & 20.61 & 16.69 & 1.11 & 1581 & 12.6 & 541 \\ 
        & \quad + \textsc{REFED} & $41.10_{\textcolor{darkgreen}{\uparrow 20.49}}$ & $40.55_{\textcolor{darkgreen}{\uparrow 23.86}}$  &  1.73  &  2069  &  $25.0_{\textcolor{darkgreen}{\uparrow 12.4}}$  &  648 \\ 

        \midrule
        & Llama-3.1-8B & - & - & - & - & - & - \\ 
        & \quad + \textsc{REFED} & 29.63 & 30.10 & 1.62 & 2095 &  12.7 & 633 \\ 

        \midrule
        & Llama-3.1-8B-Instruct & 22.90 & 23.44 & 1.49 & 2181 & 21.3 & 861 \\ 
        & \quad + \textsc{REFED} & \textbf{43.96$_{\textcolor{darkgreen}{\uparrow 21.06}}$} & \textbf{42.24$_{\textcolor{darkgreen}{\uparrow 18.80}}$} & \textbf{1.74} & \textbf{1950} & \textbf{35.9$_{\textcolor{darkgreen}{\uparrow 14.6}}$} & \textbf{670} \\ 
        
        \bottomrule
    \end{tabular}
    
    \caption{Evaluation results of finetuning \textsc{REFED} on the base and instruct variants of Llama-3.1-8B and Mistral-7B models. Green subscripts indicate improvements after finetuning. Note that we do not report base model performance because they are not instruction-tuned.}
    \label{tab:table3}
    \vspace{-15pt}
\end{table*}

\section{Experiments}

\subsection{Experimental Setup}
\paragraph{Data Synthesis.} 

We use the LIMA \citep{lima} training dataset as our seed dataset, which comprises of one thousand carefully curated instruction-response pairs. The samples were either manually written or selected from community forums, and were selected based on quality and diversity. This dataset was chosen because it is concise enough to serve as a seed dataset, while being well-designed and has demonstrated effectiveness for instruction tuning \citep{lima}.

In our experiments, we use GPT-4o mini \citep{OpenAIGpt4oMini} with our data synthesis framework to create \textsc{REFED}, an instruction tuning dataset with 10K data samples.

\paragraph{Training Setup.}

We finetune the base and instruct variants of Llama-3.1-8B \citep{grattafiori2024llama3herdmodels} and Mistral-7B \citep{jiang2023mistral7b} on \textsc{REFED}. We use a learning rate of $1 \times 10^{-6}$ for instruct variants, and $2 \times 10^{-5}$ for base variants. All other hyperparameters remain consistent across models: linear warmup ratio of 0.03, cosine decay, batch size of 128, and maximum sequence length of 2048. The models are trained for 15 epochs, with checkpoint selection based on length-controlled win-rate \citep{dubois2024lengthcontrolledalpacaevalsimpleway} on a held-out validation set of 100 synthesized instruction-response pairs that were synthesized with GPT-4o \citep{openai2024gpt4ocard}. 

When training on larger datasets like Evol Instruct \citep{xu2023wizardlmempoweringlargelanguage} and UltraChat \citep{ding-etal-2023-enhancing}, we follow prior works and modify our training setup as follows: 100 warmup steps, batch size of 32, and train for 2 epochs \citep{xu2024magpiealignmentdatasynthesis}.

\paragraph{Evaluation.}

To evaluate our model's instruction-following abilities, we use two benchmarks: AlpacaEval 2.0 \citep{dubois2024lengthcontrolledalpacaevalsimpleway} and Arena-Hard \citep{arenahard2024}. These benchmarks are automatic evaluators of language models' instruction-following abilities and have demonstrated the highest correlations with human preferences from Chatbot Arena \citep{li2024crowdsourced, dubois2024lengthcontrolledalpacaevalsimpleway}. 

Both benchmarks compute win rates by using a powerful LLM as a judge to compare model responses against established responses from a reference model. To further improve correlation with human preferences, AlpacaEval 2.0 additionally computes a length-controlled win rate that mitigates biases towards longer responses by comparing responses of similar length \citep{dubois2024lengthcontrolledalpacaevalsimpleway}.

In our experiments, we follow standard evaluation protocols and use GPT-4-Turbo (1106) as a judge. For AlpacaEval 2.0, we use GPT-4-Turbo (1106) as the reference model, and GPT-4-Turbo (0314) for the reference model in Arena-Hard.

\subsection{Experimental Results}

\subsubsection{Experiment 1: How Effective is \textsc{Reference-Level Feedback} for Data Synthesis?}

The first set of experiments evaluate the effectiveness of \textsc{Reference-Level Feedback} for data synthesis by comparing it against traditional sample-level feedback, and also systematically analyzing different components in our framework. For each approach, we synthesize datasets with 10K samples, finetune Llama-3.1-8B-Instruct on that data, then evaluate.

We conduct an ablation study by progressively introducing the different components of \textsc{Reference-Level Feedback} in our framework. Starting with a baseline of no feedback, we finetune on just our initial seed dataset. Next, we train on a dataset with 10K samples that was created by incorporating the instruction feedback and generate the corresponding response. Lastly, we evaluate our complete approach by synthesizing a dataset that also incorporates the response feedback to improve the generated response (\textsc{REFED}).

Additionally, we compare against sample-level feedback, where feedback is generated and applied individually for each response. Here, the synthesis pipeline remains consistent, with minimal prompt modifications to accommodate different feedback types. With this, we can effectively isolate the impact of different feedback strategies on response quality.

\paragraph{Results.}

Table \ref{tab:table1} shows clear improvements in performance as reference-level instruction and response feedback are introduced in the synthesis pipeline. On AlpacaEval 2.0, using the complete \textsc{Reference-Level Feedback} for data synthesis achieves a length-controlled win rate of 43.96\% and win rate of 42.25\%, showing that it is superior to sample-level feedback (LC: 42.92\%, WR: 41.74\%). Results on Arena-Hard are similar, where it achieves a win rate of 35.9\%, substantially outperforming sample-level feedback (WR: 30.8\%).

The consistent performance gains across both benchmarks demonstrate that \textsc{Reference-Level Feedback} is more effective for improving responses and generating high-quality data compared to alternative feedback types.

\subsubsection{Experiment 2: How Does Our Method Compare Against Other Baselines?} \label{sec:4.2.2}

We evaluate the performance of our synthetic data by comparing a Llama-3.1-8B-Instruct model finetuned on our dataset against several baselines.

\paragraph{Baselines.}
For the first set of baselines, we finetune Llama-3.1-8B-Instruct on various well-known synthetic datasets: Alpaca \citep{alpaca}, Evol Instruct \citep{xu2023wizardlmempoweringlargelanguage}, UltraChat 200K \citep{ding-etal-2023-enhancing}, and Instruct-Skillmix \citep{kaur2024instruct}. We use an identical training setup to the one we use for our models.

We also compare against leading models from the AlpacaEval 2.0 leaderboard that use SFT to train 8B-parameter models: Llama-3-8B-Instruct-Skillmix, which trains Llama-3-8B on the Instruct-Skillmix dataset \citep{kaur2024instruct}, and Infinity-Instruct-7M-Gen-Llama3.1-8B model, trained on Infinity-Instruct-7M and Infinity-Instruct-Gen \citep{InfinityInstruct2024}. Additionally, we consider some larger and more powerful models such as GPT-3.5, Llama-3.1-405B-Instruct \citep{dubey2024llama3herdmodels} and Claude 3 Opus \citep{anthropic_claude3}.

\paragraph{Results.}

Our results are presented in Table \ref{tab:table2}. The Llama-3.1-8B-Instruct model finetuned on \textsc{REFED} achieves state-of-the-art performance among similar sized models trained with SFT, across both evaluation benchmarks. On AlpacaEval 2.0, it achieves a length-controlled win rate of 43.96\%. This not only scores higher than our baselines, but also outperforms significantly larger models including LLama-3.1-405B-Instruct and Claude 3 Opus. On Arena-Hard, we get a win-rate of 35.9\%, outperforming both our baseline models and established models like GPT-3.5 Turbo. These results demonstrate that our data synthesis approach can enable strong model performance on established benchmarks, highlighting the effectiveness of \textsc{Reference-Level Feedback}. 

\paragraph{Human Evaluation.}
To further validate our results, we conduct a human evaluation study (see Appendix \ref{sec:human_eval}) that compares Llama-3.1-8B-Instruct finetuned on the Instruct-Skillmix dataset with Llama-3.1-8B-Instruct finetuned on \textsc{REFED}. We collected annotations from 3 independent human annotators, and use majority voting to determine final preferences. We randomly select 50 instructions for each model. The human annotators select their preferred response between a GPT-4 reference response and the model generated response.

Our human evaluations align with our results, with
Llama-3.1-8B-Instruct finetuned on the Instruct-Skillmix dataset achieving a win-rate of 44\% against GPT4, and Llama-3.1-8B-Instruct finetuned on \textsc{REFED} achieving a win-rate of 56\%.

\subsubsection{Experiment 3: Does \textsc{REFED} Generalize To Different Model Architectures?}

In this section, we evaluate the effectiveness of \textsc{REFED} across different models by finetuning both base and instruct variants of Llama-3.1-8B \citep{grattafiori2024llama3herdmodels} and Mistral-7B \citep{jiang2023mistral7b}. This analysis validates the robustness of our approach by demonstrating consistent benefits across different cases.

\paragraph{Results.}

Our results are presented in Table \ref{tab:table3}. Training on \textsc{REFED} yields improvements across all model variants. In particular, the instruct models show very strong performance. Llama-3.1-8B-Instruct-\textsc{REFED} achieves the strongest performance, with a length-controlled win rate of 43.96\% on AlpacaEval 2.0 and 35.9\% on Arena-Hard. Mistral-7B-Instruct-\textsc{REFED} shows impressive results, with 41.0\% and 25.0\% respectively. 

The base models also demonstrate notable improvements. Llama-3.1-8B-\textsc{REFED} achieves a length-controlled win rate of 29.63\% on AlpacaEval 2.0 and 12.7\% on Arena-Hard, outperforming Llama-3.1-8B-Instruct (20.9\%) on AlpacaEval 2. Similarly, Mistral-7B-\textsc{REFED} achieves 16.97\% on AlpacaEval 2.0 and 3.6\% on Arena-Hard, getting close performance to Mistral-7B-Instruct (20.7\%).

These results demonstrate that \textsc{REFED} effectively improves instruction-following capabilities across different models and model variants. The strong performance gains, particularly in base models surpassing their instruct variants, highlight how effective our dataset is in developing LLM instruction-following abilities. This observation matches the model-agnostic design of our method.

\subsubsection{Experiment 4: Does Filtering Enhance the Effectiveness?}

We explore how different filtering approaches affect model performance by finetuning Llama-3.1-8B-Instruct on various subsets of filtered data. We compare three strategies: random sampling, LLM-judge filtering, and ROUGE-L similarity filtering.
\paragraph{Random Sampling.}
As our baseline, we randomly sample subsets of size 1K, 2K, 4K, and 8K from \textsc{REFED}. 
\paragraph{LLM-Judge Filtering.}
We use GPT-4o-mini as a judge to evaluate pairs of initial and refined responses. We keep samples where refined responses are rated higher than initial responses, and obtain approximately 5K instruction-response pairs. From these, we sample subsets of size 1K, 2K, and 4K.
\paragraph{ROUGE-L Similarity Filtering.}
Following \citet{wang-etal-2023-self-instruct}, we use ROUGE-L similarity scoring to maximize instruction diversity. Starting with a randomly selected sample, we iteratively add candidates where the instruction's maximum similarity score with existing instructions is below a specific threshold. We use thresholds of 0.10, 0.11, 0.12, and 0.145 to get subsets of sizes 1K, 2K, 4K, and 8K respectively.

\begin{figure}[t]
    \centering
    \includegraphics[width=.95\linewidth]{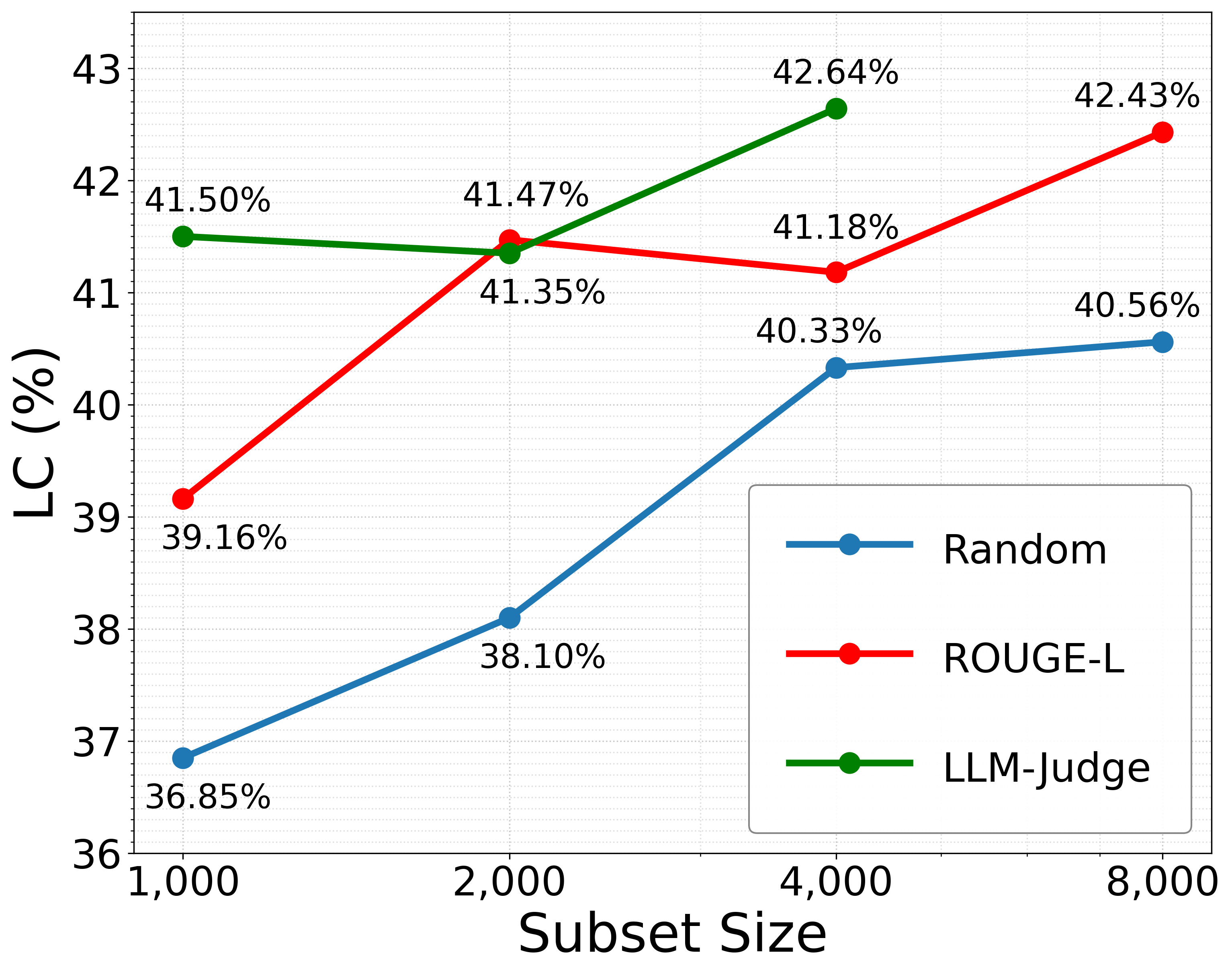} \hfill

    \caption{Length Controlled Win-Rate on AlpacaEval 2.0 for Llama-3.1-8B-Instruct finetuned on various subsets of \textsc{REFED}, based on different filtering strategies.
    }
    \label{fig:filtering}
    \vspace{-15pt}
\end{figure}

\paragraph{Analysis.} 
Our results, presented in Figure \ref{fig:filtering}, demonstrate clear benefits of filtering. LLM-Judge filtering proves most effective, achieving 42.64\% performance with just 4K samples -- comparable to the full dataset's performance with less than half the data. ROUGE-L similarity filtering performs slightly worse, but still outperforms random sampling, achieving 42.43\% with 8K samples. Although neither filtered dataset leads to higher results than the full dataset, they give comparable results while requiring less training time and computational cost. The results suggest that these filtering strategies successfully identify high-quality samples, though the slight drop in performance indicates that filtered-out responses may still contain valuable training signal.

We additionally conduct some initial experiments on scalability, and synthesize \textsc{REFED-20K}, a dataset with 20,000 samples created using our framework. The results, presented in Table \ref{tab:table5}, demonstrate significant improvements over \textsc{REFED}. These findings are promising and suggest that our framework benefits from increased scale without immediate signs of diminishing returns.

\subsection{Diversity Analysis}
A critical concern when synthesizing data is ensuring sufficient diversity. Higher diversity in synthetic training datasets has demonstrated correlation with improved LLM performance \citep{wang-etal-2023-self-instruct,yu2023large,chen2025on}. Following \citet{guo-etal-2024-curious}, we use the following metrics to analyze diversity:
\begin{itemize}[leftmargin=10pt,topsep=0pt]
\itemsep -0.5ex
     \item \textit{Type-Token Ratio}: The proportion of unique words to the total number of words.
     \item \textit{Distinct-n}: The proportion of unique n-grams to the total number of n-grams.
    \item \textit{Syntactic Diversity }: uses a neural parser to construct dependency trees for sentences, which are subsequently transformed into graphs and mapped to a vector space. Syntactic diversity is then quantified by calculating the average pairwise distance between the vector representations.
\end{itemize}

We compare our \textsc{REFED} dataset against the baseline synthetic datasets, with results presented in Table \ref{tab:diversity_metrics}. Our analysis reveals that \textsc{REFED} achieves the highest scores across the majority of diversity metrics. These results empirically validate the effectiveness of our proposed \textsc{Reference-Level Feedback} approach in synthesizing more diverse data.

\begin{table}
  \centering
  \setlength{\tabcolsep}{4pt}
  \begin{tabular}{lrrrr}
    \hline
    \textbf{Dataset} & TTR & D-2 & D-3 & SD \\
    \hline
    Alpaca & 7.66 & 51.61 & 84.40 & 61.49 \\
    Evol Instruct & 9.30 & 54.26 & 83.89 & 59.09 \\
    UltraChat & \textbf{9.42} & 57.90 & 88.36 & 61.57 \\
    Instruct-Skillmix & 7.28 & 56.14 & 87.38 & 60.67 \\
    REFED & 9.24 & \textbf{61.09} & \textbf{90.37} & \textbf{63.93} \\
    \hline
  \end{tabular}
\caption{Comparison of diversity metrics across different datasets. Type-Token Ratio (TTR), Distinct-2 (D2), Distinct-3 (D3), Syntactic Diversity (SD). Bold values indicate the highest score for each metric.}
  \label{tab:diversity_metrics}

\vspace{-15pt}
\end{table}

\subsection{Efficiency Analysis}
While \textsc{Reference-Level Feedback} involves an initial fixed cost of building a seed dataset, it is more efficient than traditional sample-level feedback in terms of the frequency of feedback collection. Sample-level approaches require feedback for every synthesized sample, $O(n)$, while our method only requires feedback a constant number of reference samples, $O(1)$.

Moreover, we achieve state-of-the-art results without requiring the most expensive language models. While approaches like \citet{kaur2024instruct} report costs of \$600 to synthesize 4K samples using GPT-4, our experiments synthesize 10K samples for less than \$20 using GPT-4o mini. Our work highlights that it is possible to synthesize high-quality data and achieve high performance, while using more economical models.

\section{Conclusion}

In this work, we introduce \textsc{Reference-Level Feedback}, a novel framework for enhancing synthetic data quality. Our approach leverages feedback collected from high-quality reference samples to identify and propagate desirable characteristics through the synthesized data. LLMs finetuned on our generated dataset, \textsc{REFED}, achieve strong performance on instruction-following benchmarks. By modeling and incorporating desirable characteristics from carefully curated reference samples, we demonstrate a powerful and efficient approach for creating high-quality instruction-tuning datasets.

Our work opens up several promising directions for future applications. Recent developments in language models like DeepSeek-R1 \citep{deepseekai2025deepseekr1incentivizingreasoningcapability} and OpenAI's o1 \citep{openai2024openaio1card} demonstrate that high-quality synthetic data plays a crucial role in achieving state-of-the-art performance. The feedback introduced in our paper can be naturally integrated with various existing data synthesis approaches to further build upon them. By demonstrating the effectiveness of \textsc{Reference-Level Feedback} in this context, this work establishes a foundation for developing more comprehensive approaches for data synthesis.
\section{Limitations}

\paragraph{Reliance on a Stronger LLM. }
Our framework relies on a stronger LLM to synthesize data. While this enables the synthesis of high quality data, removing this dependency could help lead to a more robust and independent framework, possibly at the cost of performance degradation. Additionally, LLM-generated data may amplify existing biases or include inappropriate content.

\paragraph{Seed Data Quality. }
The quality of our synthesized data is tied to that of our seed data. We select concise, high-quality datasets from prior works to use as the seed data. A more comprehensive exploration of seed data selection and its impact on synthetic data remains an important direction for future work.

Furthermore, our work does not fully address the scalability our framework. There likely exists a limit to how much data we can synthesize from our seed data, until the synthesized data becomes repetitive and lacks diversity.

\paragraph{LLM-Based Evaluation. }
Our evaluation relies on benchmarks that use LLMs as a judge. Although they correlate highly with human judgments, it is important to acknowledge that they may still have limitations, such as biases towards longer responses or their own outputs.

\section{Acknowledgments}
This work has benefited from the Microsoft Accelerate Foundation Models Research (AFMR) grant program, through which leading foundation models hosted by Microsoft Azure and access to Azure credits were provided to conduct the research.

\bibliography{custom}

\begin{thebibliography}{38}
\providecommand{\natexlab}[1]{#1}

\bibitem[{Anthropic(2025)}]{anthropic_claude3}
Anthropic. 2025.
\newblock \href {https://www.anthropic.com/news/claude-3-family} {Claude 3 family}.

\bibitem[{Bai et~al.(2022)Bai, Kadavath, Kundu, Askell, Kernion, Jones, Chen, Goldie, Mirhoseini, McKinnon, Chen, Olsson, Olah, Hernandez, Drain, Ganguli, Li, Tran-Johnson, Perez, Kerr, Mueller, Ladish, Landau, Ndousse, Lukosuite, Lovitt, Sellitto, Elhage, Schiefer, Mercado, DasSarma, Lasenby, Larson, Ringer, Johnston, Kravec, Showk, Fort, Lanham, Telleen-Lawton, Conerly, Henighan, Hume, Bowman, Hatfield-Dodds, Mann, Amodei, Joseph, McCandlish, Brown, and Kaplan}]{bai2022constitutionalaiharmlessnessai}
Yuntao Bai, Saurav Kadavath, Sandipan Kundu, Amanda Askell, Jackson Kernion, Andy Jones, Anna Chen, Anna Goldie, Azalia Mirhoseini, Cameron McKinnon, Carol Chen, Catherine Olsson, Christopher Olah, Danny Hernandez, Dawn Drain, Deep Ganguli, Dustin Li, Eli Tran-Johnson, Ethan Perez, Jamie Kerr, Jared Mueller, Jeffrey Ladish, Joshua Landau, Kamal Ndousse, Kamile Lukosuite, Liane Lovitt, Michael Sellitto, Nelson Elhage, Nicholas Schiefer, Noemi Mercado, Nova DasSarma, Robert Lasenby, Robin Larson, Sam Ringer, Scott Johnston, Shauna Kravec, Sheer~El Showk, Stanislav Fort, Tamera Lanham, Timothy Telleen-Lawton, Tom Conerly, Tom Henighan, Tristan Hume, Samuel~R. Bowman, Zac Hatfield-Dodds, Ben Mann, Dario Amodei, Nicholas Joseph, Sam McCandlish, Tom Brown, and Jared Kaplan. 2022.
\newblock \href {https://arxiv.org/abs/2212.08073} {Constitutional ai: Harmlessness from ai feedback}.
\newblock \emph{Preprint}, arXiv:2212.08073.

\bibitem[{Chen et~al.(2024{\natexlab{a}})Chen, Scheurer, Campos, Korbak, Chan, Bowman, Cho, and Perez}]{chen2024learning}
Angelica Chen, J{\'e}r{\'e}my Scheurer, Jon~Ander Campos, Tomasz Korbak, Jun~Shern Chan, Samuel~R. Bowman, Kyunghyun Cho, and Ethan Perez. 2024{\natexlab{a}}.
\newblock \href {https://openreview.net/forum?id=xo3hI5MwvU} {Learning from natural language feedback}.
\newblock \emph{Transactions on Machine Learning Research}.

\bibitem[{Chen et~al.(2025)Chen, Waheed, Li, Wang, Wang, Raj, and Abdin}]{chen2025on}
Hao Chen, Abdul Waheed, Xiang Li, Yidong Wang, Jindong Wang, Bhiksha Raj, and Marah~I Abdin. 2025.
\newblock \href {https://openreview.net/forum?id=oqsQbn4XfT} {On the diversity of synthetic data and its impact on training large language models}.

\bibitem[{Chen et~al.(2024{\natexlab{b}})Chen, Wen, Nag, Luo, Yin, Li, Li, and Wang}]{chen2024iteralign}
Xiusi Chen, Hongzhi Wen, Sreyashi Nag, Chen Luo, Qingyu Yin, Ruirui Li, Zheng Li, and Wei Wang. 2024{\natexlab{b}}.
\newblock Iteralign: Iterative constitutional alignment of large language models.
\newblock In \emph{Proceedings of the 2024 Conference of the North American Chapter of the Association for Computational Linguistics: Human Language Technologies (Volume 1: Long Papers)}, pages 1423--1433.

\bibitem[{DeepSeek-AI et~al.(2025)DeepSeek-AI, Guo, Yang, Zhang, Song, Zhang, Xu, Zhu, Ma, Wang, Bi, Zhang, Yu, Wu, Wu, Gou, Shao, Li, Gao, Liu, Xue, Wang, Wu, Feng, Lu, Zhao, Deng, Zhang, Ruan, Dai, Chen, Ji, Li, Lin, Dai, Luo, Hao, Chen, Li, Zhang, Bao, Xu, Wang, Ding, Xin, Gao, Qu, Li, Guo, Li, Wang, Chen, Yuan, Qiu, Li, Cai, Ni, Liang, Chen, Dong, Hu, Gao, Guan, Huang, Yu, Wang, Zhang, Zhao, Wang, Zhang, Xu, Xia, Zhang, Zhang, Tang, Li, Wang, Li, Tian, Huang, Zhang, Wang, Chen, Du, Ge, Zhang, Pan, Wang, Chen, Jin, Chen, Lu, Zhou, Chen, Ye, Wang, Yu, Zhou, Pan, Li, Zhou, Wu, Ye, Yun, Pei, Sun, Wang, Zeng, Zhao, Liu, Liang, Gao, Yu, Zhang, Xiao, An, Liu, Wang, Chen, Nie, Cheng, Liu, Xie, Liu, Yang, Li, Su, Lin, Li, Jin, Shen, Chen, Sun, Wang, Song, Zhou, Wang, Shan, Li, Wang, Wei, Zhang, Xu, Li, Zhao, Sun, Wang, Yu, Zhang, Shi, Xiong, He, Piao, Wang, Tan, Ma, Liu, Guo, Ou, Wang, Gong, Zou, He, Xiong, Luo, You, Liu, Zhou, Zhu, Xu, Huang, Li, Zheng, Zhu, Ma, Tang, Zha, Yan, Ren, Ren, Sha, Fu, Xu, Xie, Zhang,
  Hao, Ma, Yan, Wu, Gu, Zhu, Liu, Li, Xie, Song, Pan, Huang, Xu, Zhang, and Zhang}]{deepseekai2025deepseekr1incentivizingreasoningcapability}
DeepSeek-AI, Daya Guo, Dejian Yang, Haowei Zhang, Junxiao Song, Ruoyu Zhang, Runxin Xu, Qihao Zhu, Shirong Ma, Peiyi Wang, Xiao Bi, Xiaokang Zhang, Xingkai Yu, Yu~Wu, Z.~F. Wu, Zhibin Gou, Zhihong Shao, Zhuoshu Li, Ziyi Gao, Aixin Liu, Bing Xue, Bingxuan Wang, Bochao Wu, Bei Feng, Chengda Lu, Chenggang Zhao, Chengqi Deng, Chenyu Zhang, Chong Ruan, Damai Dai, Deli Chen, Dongjie Ji, Erhang Li, Fangyun Lin, Fucong Dai, Fuli Luo, Guangbo Hao, Guanting Chen, Guowei Li, H.~Zhang, Han Bao, Hanwei Xu, Haocheng Wang, Honghui Ding, Huajian Xin, Huazuo Gao, Hui Qu, Hui Li, Jianzhong Guo, Jiashi Li, Jiawei Wang, Jingchang Chen, Jingyang Yuan, Junjie Qiu, Junlong Li, J.~L. Cai, Jiaqi Ni, Jian Liang, Jin Chen, Kai Dong, Kai Hu, Kaige Gao, Kang Guan, Kexin Huang, Kuai Yu, Lean Wang, Lecong Zhang, Liang Zhao, Litong Wang, Liyue Zhang, Lei Xu, Leyi Xia, Mingchuan Zhang, Minghua Zhang, Minghui Tang, Meng Li, Miaojun Wang, Mingming Li, Ning Tian, Panpan Huang, Peng Zhang, Qiancheng Wang, Qinyu Chen, Qiushi Du, Ruiqi Ge, Ruisong
  Zhang, Ruizhe Pan, Runji Wang, R.~J. Chen, R.~L. Jin, Ruyi Chen, Shanghao Lu, Shangyan Zhou, Shanhuang Chen, Shengfeng Ye, Shiyu Wang, Shuiping Yu, Shunfeng Zhou, Shuting Pan, S.~S. Li, Shuang Zhou, Shaoqing Wu, Shengfeng Ye, Tao Yun, Tian Pei, Tianyu Sun, T.~Wang, Wangding Zeng, Wanjia Zhao, Wen Liu, Wenfeng Liang, Wenjun Gao, Wenqin Yu, Wentao Zhang, W.~L. Xiao, Wei An, Xiaodong Liu, Xiaohan Wang, Xiaokang Chen, Xiaotao Nie, Xin Cheng, Xin Liu, Xin Xie, Xingchao Liu, Xinyu Yang, Xinyuan Li, Xuecheng Su, Xuheng Lin, X.~Q. Li, Xiangyue Jin, Xiaojin Shen, Xiaosha Chen, Xiaowen Sun, Xiaoxiang Wang, Xinnan Song, Xinyi Zhou, Xianzu Wang, Xinxia Shan, Y.~K. Li, Y.~Q. Wang, Y.~X. Wei, Yang Zhang, Yanhong Xu, Yao Li, Yao Zhao, Yaofeng Sun, Yaohui Wang, Yi~Yu, Yichao Zhang, Yifan Shi, Yiliang Xiong, Ying He, Yishi Piao, Yisong Wang, Yixuan Tan, Yiyang Ma, Yiyuan Liu, Yongqiang Guo, Yuan Ou, Yuduan Wang, Yue Gong, Yuheng Zou, Yujia He, Yunfan Xiong, Yuxiang Luo, Yuxiang You, Yuxuan Liu, Yuyang Zhou, Y.~X. Zhu,
  Yanhong Xu, Yanping Huang, Yaohui Li, Yi~Zheng, Yuchen Zhu, Yunxian Ma, Ying Tang, Yukun Zha, Yuting Yan, Z.~Z. Ren, Zehui Ren, Zhangli Sha, Zhe Fu, Zhean Xu, Zhenda Xie, Zhengyan Zhang, Zhewen Hao, Zhicheng Ma, Zhigang Yan, Zhiyu Wu, Zihui Gu, Zijia Zhu, Zijun Liu, Zilin Li, Ziwei Xie, Ziyang Song, Zizheng Pan, Zhen Huang, Zhipeng Xu, Zhongyu Zhang, and Zhen Zhang. 2025.
\newblock \href {https://arxiv.org/abs/2501.12948} {Deepseek-r1: Incentivizing reasoning capability in llms via reinforcement learning}.
\newblock \emph{Preprint}, arXiv:2501.12948.

\bibitem[{Ding et~al.(2023)Ding, Chen, Xu, Qin, Hu, Liu, Sun, and Zhou}]{ding-etal-2023-enhancing}
Ning Ding, Yulin Chen, Bokai Xu, Yujia Qin, Shengding Hu, Zhiyuan Liu, Maosong Sun, and Bowen Zhou. 2023.
\newblock \href {https://doi.org/10.18653/v1/2023.emnlp-main.183} {Enhancing chat language models by scaling high-quality instructional conversations}.
\newblock pages 3029--3051, Singapore.

\bibitem[{Dubey et~al.(2024)Dubey, Jauhri, Pandey, Kadian, Al-Dahle, Letman, Mathur, Schelten, Yang, Fan, Goyal, Hartshorn, Yang, Mitra, Sravankumar, Korenev, Hinsvark, Rao, Zhang, Rodriguez, Gregerson, Spataru, Roziere, Biron, Tang, Chern, Caucheteux, Nayak, Bi, Marra, McConnell, Keller, Touret, Wu, Wong, Ferrer, Nikolaidis, Allonsius, Song, Pintz, Livshits, Esiobu, Choudhary, Mahajan, Garcia-Olano, Perino, Hupkes, Lakomkin, AlBadawy, Lobanova, Dinan, Smith, Radenovic, Zhang, Synnaeve, Lee, Anderson, Nail, Mialon, Pang, Cucurell, Nguyen, Korevaar, Xu, Touvron, Zarov, Ibarra, Kloumann, Misra, Evtimov, Copet, Lee, Geffert, Vranes, Park, Mahadeokar, Shah, van~der Linde, Billock, Hong, Lee, Fu, Chi, Huang, Liu, Wang, Yu, Bitton, Spisak, Park, Rocca, Johnstun, Saxe, Jia, Alwala, Upasani, Plawiak, Li, Heafield, Stone, El-Arini, Iyer, Malik, Chiu, Bhalla, Rantala-Yeary, van~der Maaten, Chen, Tan, Jenkins, Martin, Madaan, Malo, Blecher, Landzaat, de~Oliveira, Muzzi, Pasupuleti, Singh, Paluri, Kardas, Oldham, Rita,
  Pavlova, Kambadur, Lewis, Si, Singh, Hassan, Goyal, Torabi, Bashlykov, Bogoychev, Chatterji, Duchenne, Çelebi, Alrassy, Zhang, Li, Vasic, Weng, Bhargava, Dubal, Krishnan, Koura, Xu, He, Dong, Srinivasan, Ganapathy, Calderer, Cabral, Stojnic, Raileanu, Girdhar, Patel, Sauvestre, Polidoro, Sumbaly, Taylor, Silva, Hou, Wang, Hosseini, Chennabasappa, Singh, Bell, Kim, Edunov, Nie, Narang, Raparthy, Shen, Wan, Bhosale, Zhang, Vandenhende, Batra, Whitman, Sootla, Collot, Gururangan, Borodinsky, Herman, Fowler, Sheasha, Georgiou, Scialom, Speckbacher, Mihaylov, Xiao, Karn, Goswami, Gupta, Ramanathan, Kerkez, Gonguet, Do, Vogeti, Petrovic, Chu, Xiong, Fu, Meers, Martinet, Wang, Tan, Xie, Jia, Wang, Goldschlag, Gaur, Babaei, Wen, Song, Zhang, Li, Mao, Coudert, Yan, Chen, Papakipos, Singh, Grattafiori, Jain, Kelsey, Shajnfeld, Gangidi, Victoria, Goldstand, Menon, Sharma, Boesenberg, Vaughan, Baevski, Feinstein, Kallet, Sangani, Yunus, Lupu, Alvarado, Caples, Gu, Ho, Poulton, Ryan, Ramchandani, Franco, Saraf,
  Chowdhury, Gabriel, Bharambe, Eisenman, Yazdan, James, Maurer, Leonhardi, Huang, Loyd, Paola, Paranjape, Liu, Wu, Ni, Hancock, Wasti, Spence, Stojkovic, Gamido, Montalvo, Parker, Burton, Mejia, Wang, Kim, Zhou, Hu, Chu, Cai, Tindal, Feichtenhofer, Civin, Beaty, Kreymer, Li, Wyatt, Adkins, Xu, Testuggine, David, Parikh, Liskovich, Foss, Wang, Le, Holland, Dowling, Jamil, Montgomery, Presani, Hahn, Wood, Brinkman, Arcaute, Dunbar, Smothers, Sun, Kreuk, Tian, Ozgenel, Caggioni, Guzmán, Kanayet, Seide, Florez, Schwarz, Badeer, Swee, Halpern, Thattai, Herman, Sizov, Guangyi, Zhang, Lakshminarayanan, Shojanazeri, Zou, Wang, Zha, Habeeb, Rudolph, Suk, Aspegren, Goldman, Damlaj, Molybog, Tufanov, Veliche, Gat, Weissman, Geboski, Kohli, Asher, Gaya, Marcus, Tang, Chan, Zhen, Reizenstein, Teboul, Zhong, Jin, Yang, Cummings, Carvill, Shepard, McPhie, Torres, Ginsburg, Wang, Wu, U, Saxena, Prasad, Khandelwal, Zand, Matosich, Veeraraghavan, Michelena, Li, Huang, Chawla, Lakhotia, Huang, Chen, Garg, A, Silva, Bell,
  Zhang, Guo, Yu, Moshkovich, Wehrstedt, Khabsa, Avalani, Bhatt, Tsimpoukelli, Mankus, Hasson, Lennie, Reso, Groshev, Naumov, Lathi, Keneally, Seltzer, Valko, Restrepo, Patel, Vyatskov, Samvelyan, Clark, Macey, Wang, Hermoso, Metanat, Rastegari, Bansal, Santhanam, Parks, White, Bawa, Singhal, Egebo, Usunier, Laptev, Dong, Zhang, Cheng, Chernoguz, Hart, Salpekar, Kalinli, Kent, Parekh, Saab, Balaji, Rittner, Bontrager, Roux, Dollar, Zvyagina, Ratanchandani, Yuvraj, Liang, Alao, Rodriguez, Ayub, Murthy, Nayani, Mitra, Li, Hogan, Battey, Wang, Maheswari, Howes, Rinott, Bondu, Datta, Chugh, Hunt, Dhillon, Sidorov, Pan, Verma, Yamamoto, Ramaswamy, Lindsay, Lindsay, Feng, Lin, Zha, Shankar, Zhang, Zhang, Wang, Agarwal, Sajuyigbe, Chintala, Max, Chen, Kehoe, Satterfield, Govindaprasad, Gupta, Cho, Virk, Subramanian, Choudhury, Goldman, Remez, Glaser, Best, Kohler, Robinson, Li, Zhang, Matthews, Chou, Shaked, Vontimitta, Ajayi, Montanez, Mohan, Kumar, Mangla, Albiero, Ionescu, Poenaru, Mihailescu, Ivanov, Li, Wang,
  Jiang, Bouaziz, Constable, Tang, Wang, Wu, Wang, Xia, Wu, Gao, Chen, Hu, Jia, Qi, Li, Zhang, Zhang, Adi, Nam, Yu, Wang, Hao, Qian, He, Rait, DeVito, Rosnbrick, Wen, Yang, and Zhao}]{dubey2024llama3herdmodels}
Abhimanyu Dubey, Abhinav Jauhri, Abhinav Pandey, Abhishek Kadian, Ahmad Al-Dahle, Aiesha Letman, Akhil Mathur, Alan Schelten, Amy Yang, Angela Fan, Anirudh Goyal, Anthony Hartshorn, Aobo Yang, Archi Mitra, Archie Sravankumar, Artem Korenev, Arthur Hinsvark, Arun Rao, Aston Zhang, Aurelien Rodriguez, Austen Gregerson, Ava Spataru, Baptiste Roziere, Bethany Biron, Binh Tang, Bobbie Chern, Charlotte Caucheteux, Chaya Nayak, Chloe Bi, Chris Marra, Chris McConnell, Christian Keller, Christophe Touret, Chunyang Wu, Corinne Wong, Cristian~Canton Ferrer, Cyrus Nikolaidis, Damien Allonsius, Daniel Song, Danielle Pintz, Danny Livshits, David Esiobu, Dhruv Choudhary, Dhruv Mahajan, Diego Garcia-Olano, Diego Perino, Dieuwke Hupkes, Egor Lakomkin, Ehab AlBadawy, Elina Lobanova, Emily Dinan, Eric~Michael Smith, Filip Radenovic, Frank Zhang, Gabriel Synnaeve, Gabrielle Lee, Georgia~Lewis Anderson, Graeme Nail, Gregoire Mialon, Guan Pang, Guillem Cucurell, Hailey Nguyen, Hannah Korevaar, Hu~Xu, Hugo Touvron, Iliyan Zarov,
  Imanol~Arrieta Ibarra, Isabel Kloumann, Ishan Misra, Ivan Evtimov, Jade Copet, Jaewon Lee, Jan Geffert, Jana Vranes, Jason Park, Jay Mahadeokar, Jeet Shah, Jelmer van~der Linde, Jennifer Billock, Jenny Hong, Jenya Lee, Jeremy Fu, Jianfeng Chi, Jianyu Huang, Jiawen Liu, Jie Wang, Jiecao Yu, Joanna Bitton, Joe Spisak, Jongsoo Park, Joseph Rocca, Joshua Johnstun, Joshua Saxe, Junteng Jia, Kalyan~Vasuden Alwala, Kartikeya Upasani, Kate Plawiak, Ke~Li, Kenneth Heafield, Kevin Stone, Khalid El-Arini, Krithika Iyer, Kshitiz Malik, Kuenley Chiu, Kunal Bhalla, Lauren Rantala-Yeary, Laurens van~der Maaten, Lawrence Chen, Liang Tan, Liz Jenkins, Louis Martin, Lovish Madaan, Lubo Malo, Lukas Blecher, Lukas Landzaat, Luke de~Oliveira, Madeline Muzzi, Mahesh Pasupuleti, Mannat Singh, Manohar Paluri, Marcin Kardas, Mathew Oldham, Mathieu Rita, Maya Pavlova, Melanie Kambadur, Mike Lewis, Min Si, Mitesh~Kumar Singh, Mona Hassan, Naman Goyal, Narjes Torabi, Nikolay Bashlykov, Nikolay Bogoychev, Niladri Chatterji, Olivier
  Duchenne, Onur Çelebi, Patrick Alrassy, Pengchuan Zhang, Pengwei Li, Petar Vasic, Peter Weng, Prajjwal Bhargava, Pratik Dubal, Praveen Krishnan, Punit~Singh Koura, Puxin Xu, Qing He, Qingxiao Dong, Ragavan Srinivasan, Raj Ganapathy, Ramon Calderer, Ricardo~Silveira Cabral, Robert Stojnic, Roberta Raileanu, Rohit Girdhar, Rohit Patel, Romain Sauvestre, Ronnie Polidoro, Roshan Sumbaly, Ross Taylor, Ruan Silva, Rui Hou, Rui Wang, Saghar Hosseini, Sahana Chennabasappa, Sanjay Singh, Sean Bell, Seohyun~Sonia Kim, Sergey Edunov, Shaoliang Nie, Sharan Narang, Sharath Raparthy, Sheng Shen, Shengye Wan, Shruti Bhosale, Shun Zhang, Simon Vandenhende, Soumya Batra, Spencer Whitman, Sten Sootla, Stephane Collot, Suchin Gururangan, Sydney Borodinsky, Tamar Herman, Tara Fowler, Tarek Sheasha, Thomas Georgiou, Thomas Scialom, Tobias Speckbacher, Todor Mihaylov, Tong Xiao, Ujjwal Karn, Vedanuj Goswami, Vibhor Gupta, Vignesh Ramanathan, Viktor Kerkez, Vincent Gonguet, Virginie Do, Vish Vogeti, Vladan Petrovic, Weiwei Chu,
  Wenhan Xiong, Wenyin Fu, Whitney Meers, Xavier Martinet, Xiaodong Wang, Xiaoqing~Ellen Tan, Xinfeng Xie, Xuchao Jia, Xuewei Wang, Yaelle Goldschlag, Yashesh Gaur, Yasmine Babaei, Yi~Wen, Yiwen Song, Yuchen Zhang, Yue Li, Yuning Mao, Zacharie~Delpierre Coudert, Zheng Yan, Zhengxing Chen, Zoe Papakipos, Aaditya Singh, Aaron Grattafiori, Abha Jain, Adam Kelsey, Adam Shajnfeld, Adithya Gangidi, Adolfo Victoria, Ahuva Goldstand, Ajay Menon, Ajay Sharma, Alex Boesenberg, Alex Vaughan, Alexei Baevski, Allie Feinstein, Amanda Kallet, Amit Sangani, Anam Yunus, Andrei Lupu, Andres Alvarado, Andrew Caples, Andrew Gu, Andrew Ho, Andrew Poulton, Andrew Ryan, Ankit Ramchandani, Annie Franco, Aparajita Saraf, Arkabandhu Chowdhury, Ashley Gabriel, Ashwin Bharambe, Assaf Eisenman, Azadeh Yazdan, Beau James, Ben Maurer, Benjamin Leonhardi, Bernie Huang, Beth Loyd, Beto~De Paola, Bhargavi Paranjape, Bing Liu, Bo~Wu, Boyu Ni, Braden Hancock, Bram Wasti, Brandon Spence, Brani Stojkovic, Brian Gamido, Britt Montalvo, Carl
  Parker, Carly Burton, Catalina Mejia, Changhan Wang, Changkyu Kim, Chao Zhou, Chester Hu, Ching-Hsiang Chu, Chris Cai, Chris Tindal, Christoph Feichtenhofer, Damon Civin, Dana Beaty, Daniel Kreymer, Daniel Li, Danny Wyatt, David Adkins, David Xu, Davide Testuggine, Delia David, Devi Parikh, Diana Liskovich, Didem Foss, Dingkang Wang, Duc Le, Dustin Holland, Edward Dowling, Eissa Jamil, Elaine Montgomery, Eleonora Presani, Emily Hahn, Emily Wood, Erik Brinkman, Esteban Arcaute, Evan Dunbar, Evan Smothers, Fei Sun, Felix Kreuk, Feng Tian, Firat Ozgenel, Francesco Caggioni, Francisco Guzmán, Frank Kanayet, Frank Seide, Gabriela~Medina Florez, Gabriella Schwarz, Gada Badeer, Georgia Swee, Gil Halpern, Govind Thattai, Grant Herman, Grigory Sizov, Guangyi, Zhang, Guna Lakshminarayanan, Hamid Shojanazeri, Han Zou, Hannah Wang, Hanwen Zha, Haroun Habeeb, Harrison Rudolph, Helen Suk, Henry Aspegren, Hunter Goldman, Ibrahim Damlaj, Igor Molybog, Igor Tufanov, Irina-Elena Veliche, Itai Gat, Jake Weissman, James
  Geboski, James Kohli, Japhet Asher, Jean-Baptiste Gaya, Jeff Marcus, Jeff Tang, Jennifer Chan, Jenny Zhen, Jeremy Reizenstein, Jeremy Teboul, Jessica Zhong, Jian Jin, Jingyi Yang, Joe Cummings, Jon Carvill, Jon Shepard, Jonathan McPhie, Jonathan Torres, Josh Ginsburg, Junjie Wang, Kai Wu, Kam~Hou U, Karan Saxena, Karthik Prasad, Kartikay Khandelwal, Katayoun Zand, Kathy Matosich, Kaushik Veeraraghavan, Kelly Michelena, Keqian Li, Kun Huang, Kunal Chawla, Kushal Lakhotia, Kyle Huang, Lailin Chen, Lakshya Garg, Lavender A, Leandro Silva, Lee Bell, Lei Zhang, Liangpeng Guo, Licheng Yu, Liron Moshkovich, Luca Wehrstedt, Madian Khabsa, Manav Avalani, Manish Bhatt, Maria Tsimpoukelli, Martynas Mankus, Matan Hasson, Matthew Lennie, Matthias Reso, Maxim Groshev, Maxim Naumov, Maya Lathi, Meghan Keneally, Michael~L. Seltzer, Michal Valko, Michelle Restrepo, Mihir Patel, Mik Vyatskov, Mikayel Samvelyan, Mike Clark, Mike Macey, Mike Wang, Miquel~Jubert Hermoso, Mo~Metanat, Mohammad Rastegari, Munish Bansal, Nandhini
  Santhanam, Natascha Parks, Natasha White, Navyata Bawa, Nayan Singhal, Nick Egebo, Nicolas Usunier, Nikolay~Pavlovich Laptev, Ning Dong, Ning Zhang, Norman Cheng, Oleg Chernoguz, Olivia Hart, Omkar Salpekar, Ozlem Kalinli, Parkin Kent, Parth Parekh, Paul Saab, Pavan Balaji, Pedro Rittner, Philip Bontrager, Pierre Roux, Piotr Dollar, Polina Zvyagina, Prashant Ratanchandani, Pritish Yuvraj, Qian Liang, Rachad Alao, Rachel Rodriguez, Rafi Ayub, Raghotham Murthy, Raghu Nayani, Rahul Mitra, Raymond Li, Rebekkah Hogan, Robin Battey, Rocky Wang, Rohan Maheswari, Russ Howes, Ruty Rinott, Sai~Jayesh Bondu, Samyak Datta, Sara Chugh, Sara Hunt, Sargun Dhillon, Sasha Sidorov, Satadru Pan, Saurabh Verma, Seiji Yamamoto, Sharadh Ramaswamy, Shaun Lindsay, Shaun Lindsay, Sheng Feng, Shenghao Lin, Shengxin~Cindy Zha, Shiva Shankar, Shuqiang Zhang, Shuqiang Zhang, Sinong Wang, Sneha Agarwal, Soji Sajuyigbe, Soumith Chintala, Stephanie Max, Stephen Chen, Steve Kehoe, Steve Satterfield, Sudarshan Govindaprasad, Sumit Gupta,
  Sungmin Cho, Sunny Virk, Suraj Subramanian, Sy~Choudhury, Sydney Goldman, Tal Remez, Tamar Glaser, Tamara Best, Thilo Kohler, Thomas Robinson, Tianhe Li, Tianjun Zhang, Tim Matthews, Timothy Chou, Tzook Shaked, Varun Vontimitta, Victoria Ajayi, Victoria Montanez, Vijai Mohan, Vinay~Satish Kumar, Vishal Mangla, Vítor Albiero, Vlad Ionescu, Vlad Poenaru, Vlad~Tiberiu Mihailescu, Vladimir Ivanov, Wei Li, Wenchen Wang, Wenwen Jiang, Wes Bouaziz, Will Constable, Xiaocheng Tang, Xiaofang Wang, Xiaojian Wu, Xiaolan Wang, Xide Xia, Xilun Wu, Xinbo Gao, Yanjun Chen, Ye~Hu, Ye~Jia, Ye~Qi, Yenda Li, Yilin Zhang, Ying Zhang, Yossi Adi, Youngjin Nam, Yu, Wang, Yuchen Hao, Yundi Qian, Yuzi He, Zach Rait, Zachary DeVito, Zef Rosnbrick, Zhaoduo Wen, Zhenyu Yang, and Zhiwei Zhao. 2024.
\newblock \href {https://arxiv.org/abs/2407.21783} {The llama 3 herd of models}.
\newblock \emph{Preprint}, arXiv:2407.21783.

\bibitem[{Dubois et~al.(2024)Dubois, Galambosi, Liang, and Hashimoto}]{dubois2024lengthcontrolledalpacaevalsimpleway}
Yann Dubois, Balázs Galambosi, Percy Liang, and Tatsunori~B. Hashimoto. 2024.
\newblock \href {https://arxiv.org/abs/2404.04475} {Length-controlled alpacaeval: A simple way to debias automatic evaluators}.
\newblock \emph{Preprint}, arXiv:2404.04475.

\bibitem[{Gou et~al.(2024)Gou, Shao, Gong, Shen, Yang, Duan, and Chen}]{gou2024criticlargelanguagemodels}
Zhibin Gou, Zhihong Shao, Yeyun Gong, Yelong Shen, Yujiu Yang, Nan Duan, and Weizhu Chen. 2024.
\newblock \href {https://arxiv.org/abs/2305.11738} {Critic: Large language models can self-correct with tool-interactive critiquing}.
\newblock \emph{Preprint}, arXiv:2305.11738.

\bibitem[{Grattafiori et~al.(2024)Grattafiori, Dubey, Jauhri, Pandey, Kadian, Al-Dahle, Letman, Mathur, Schelten, Vaughan, Yang, Fan, Goyal, Hartshorn, Yang, Mitra, Sravankumar, Korenev, Hinsvark, Rao, Zhang, Rodriguez, Gregerson, Spataru, Roziere, Biron, Tang, Chern, Caucheteux, Nayak, Bi, Marra, McConnell, Keller, Touret, Wu, Wong, Ferrer, Nikolaidis, Allonsius, Song, Pintz, Livshits, Wyatt, Esiobu, Choudhary, Mahajan, Garcia-Olano, Perino, Hupkes, Lakomkin, AlBadawy, Lobanova, Dinan, Smith, Radenovic, Guzmán, Zhang, Synnaeve, Lee, Anderson, Thattai, Nail, Mialon, Pang, Cucurell, Nguyen, Korevaar, Xu, Touvron, Zarov, Ibarra, Kloumann, Misra, Evtimov, Zhang, Copet, Lee, Geffert, Vranes, Park, Mahadeokar, Shah, van~der Linde, Billock, Hong, Lee, Fu, Chi, Huang, Liu, Wang, Yu, Bitton, Spisak, Park, Rocca, Johnstun, Saxe, Jia, Alwala, Prasad, Upasani, Plawiak, Li, Heafield, Stone, El-Arini, Iyer, Malik, Chiu, Bhalla, Lakhotia, Rantala-Yeary, van~der Maaten, Chen, Tan, Jenkins, Martin, Madaan, Malo, Blecher,
  Landzaat, de~Oliveira, Muzzi, Pasupuleti, Singh, Paluri, Kardas, Tsimpoukelli, Oldham, Rita, Pavlova, Kambadur, Lewis, Si, Singh, Hassan, Goyal, Torabi, Bashlykov, Bogoychev, Chatterji, Zhang, Duchenne, Çelebi, Alrassy, Zhang, Li, Vasic, Weng, Bhargava, Dubal, Krishnan, Koura, Xu, He, Dong, Srinivasan, Ganapathy, Calderer, Cabral, Stojnic, Raileanu, Maheswari, Girdhar, Patel, Sauvestre, Polidoro, Sumbaly, Taylor, Silva, Hou, Wang, Hosseini, Chennabasappa, Singh, Bell, Kim, Edunov, Nie, Narang, Raparthy, Shen, Wan, Bhosale, Zhang, Vandenhende, Batra, Whitman, Sootla, Collot, Gururangan, Borodinsky, Herman, Fowler, Sheasha, Georgiou, Scialom, Speckbacher, Mihaylov, Xiao, Karn, Goswami, Gupta, Ramanathan, Kerkez, Gonguet, Do, Vogeti, Albiero, Petrovic, Chu, Xiong, Fu, Meers, Martinet, Wang, Wang, Tan, Xia, Xie, Jia, Wang, Goldschlag, Gaur, Babaei, Wen, Song, Zhang, Li, Mao, Coudert, Yan, Chen, Papakipos, Singh, Srivastava, Jain, Kelsey, Shajnfeld, Gangidi, Victoria, Goldstand, Menon, Sharma, Boesenberg,
  Baevski, Feinstein, Kallet, Sangani, Teo, Yunus, Lupu, Alvarado, Caples, Gu, Ho, Poulton, Ryan, Ramchandani, Dong, Franco, Goyal, Saraf, Chowdhury, Gabriel, Bharambe, Eisenman, Yazdan, James, Maurer, Leonhardi, Huang, Loyd, Paola, Paranjape, Liu, Wu, Ni, Hancock, Wasti, Spence, Stojkovic, Gamido, Montalvo, Parker, Burton, Mejia, Liu, Wang, Kim, Zhou, Hu, Chu, Cai, Tindal, Feichtenhofer, Gao, Civin, Beaty, Kreymer, Li, Adkins, Xu, Testuggine, David, Parikh, Liskovich, Foss, Wang, Le, Holland, Dowling, Jamil, Montgomery, Presani, Hahn, Wood, Le, Brinkman, Arcaute, Dunbar, Smothers, Sun, Kreuk, Tian, Kokkinos, Ozgenel, Caggioni, Kanayet, Seide, Florez, Schwarz, Badeer, Swee, Halpern, Herman, Sizov, Guangyi, Zhang, Lakshminarayanan, Inan, Shojanazeri, Zou, Wang, Zha, Habeeb, Rudolph, Suk, Aspegren, Goldman, Zhan, Damlaj, Molybog, Tufanov, Leontiadis, Veliche, Gat, Weissman, Geboski, Kohli, Lam, Asher, Gaya, Marcus, Tang, Chan, Zhen, Reizenstein, Teboul, Zhong, Jin, Yang, Cummings, Carvill, Shepard, McPhie,
  Torres, Ginsburg, Wang, Wu, U, Saxena, Khandelwal, Zand, Matosich, Veeraraghavan, Michelena, Li, Jagadeesh, Huang, Chawla, Huang, Chen, Garg, A, Silva, Bell, Zhang, Guo, Yu, Moshkovich, Wehrstedt, Khabsa, Avalani, Bhatt, Mankus, Hasson, Lennie, Reso, Groshev, Naumov, Lathi, Keneally, Liu, Seltzer, Valko, Restrepo, Patel, Vyatskov, Samvelyan, Clark, Macey, Wang, Hermoso, Metanat, Rastegari, Bansal, Santhanam, Parks, White, Bawa, Singhal, Egebo, Usunier, Mehta, Laptev, Dong, Cheng, Chernoguz, Hart, Salpekar, Kalinli, Kent, Parekh, Saab, Balaji, Rittner, Bontrager, Roux, Dollar, Zvyagina, Ratanchandani, Yuvraj, Liang, Alao, Rodriguez, Ayub, Murthy, Nayani, Mitra, Parthasarathy, Li, Hogan, Battey, Wang, Howes, Rinott, Mehta, Siby, Bondu, Datta, Chugh, Hunt, Dhillon, Sidorov, Pan, Mahajan, Verma, Yamamoto, Ramaswamy, Lindsay, Lindsay, Feng, Lin, Zha, Patil, Shankar, Zhang, Zhang, Wang, Agarwal, Sajuyigbe, Chintala, Max, Chen, Kehoe, Satterfield, Govindaprasad, Gupta, Deng, Cho, Virk, Subramanian, Choudhury,
  Goldman, Remez, Glaser, Best, Koehler, Robinson, Li, Zhang, Matthews, Chou, Shaked, Vontimitta, Ajayi, Montanez, Mohan, Kumar, Mangla, Ionescu, Poenaru, Mihailescu, Ivanov, Li, Wang, Jiang, Bouaziz, Constable, Tang, Wu, Wang, Wu, Gao, Kleinman, Chen, Hu, Jia, Qi, Li, Zhang, Zhang, Adi, Nam, Yu, Wang, Zhao, Hao, Qian, Li, He, Rait, DeVito, Rosnbrick, Wen, Yang, Zhao, and Ma}]{grattafiori2024llama3herdmodels}
Aaron Grattafiori, Abhimanyu Dubey, Abhinav Jauhri, Abhinav Pandey, Abhishek Kadian, Ahmad Al-Dahle, Aiesha Letman, Akhil Mathur, Alan Schelten, Alex Vaughan, Amy Yang, Angela Fan, Anirudh Goyal, Anthony Hartshorn, Aobo Yang, Archi Mitra, Archie Sravankumar, Artem Korenev, Arthur Hinsvark, Arun Rao, Aston Zhang, Aurelien Rodriguez, Austen Gregerson, Ava Spataru, Baptiste Roziere, Bethany Biron, Binh Tang, Bobbie Chern, Charlotte Caucheteux, Chaya Nayak, Chloe Bi, Chris Marra, Chris McConnell, Christian Keller, Christophe Touret, Chunyang Wu, Corinne Wong, Cristian~Canton Ferrer, Cyrus Nikolaidis, Damien Allonsius, Daniel Song, Danielle Pintz, Danny Livshits, Danny Wyatt, David Esiobu, Dhruv Choudhary, Dhruv Mahajan, Diego Garcia-Olano, Diego Perino, Dieuwke Hupkes, Egor Lakomkin, Ehab AlBadawy, Elina Lobanova, Emily Dinan, Eric~Michael Smith, Filip Radenovic, Francisco Guzmán, Frank Zhang, Gabriel Synnaeve, Gabrielle Lee, Georgia~Lewis Anderson, Govind Thattai, Graeme Nail, Gregoire Mialon, Guan Pang,
  Guillem Cucurell, Hailey Nguyen, Hannah Korevaar, Hu~Xu, Hugo Touvron, Iliyan Zarov, Imanol~Arrieta Ibarra, Isabel Kloumann, Ishan Misra, Ivan Evtimov, Jack Zhang, Jade Copet, Jaewon Lee, Jan Geffert, Jana Vranes, Jason Park, Jay Mahadeokar, Jeet Shah, Jelmer van~der Linde, Jennifer Billock, Jenny Hong, Jenya Lee, Jeremy Fu, Jianfeng Chi, Jianyu Huang, Jiawen Liu, Jie Wang, Jiecao Yu, Joanna Bitton, Joe Spisak, Jongsoo Park, Joseph Rocca, Joshua Johnstun, Joshua Saxe, Junteng Jia, Kalyan~Vasuden Alwala, Karthik Prasad, Kartikeya Upasani, Kate Plawiak, Ke~Li, Kenneth Heafield, Kevin Stone, Khalid El-Arini, Krithika Iyer, Kshitiz Malik, Kuenley Chiu, Kunal Bhalla, Kushal Lakhotia, Lauren Rantala-Yeary, Laurens van~der Maaten, Lawrence Chen, Liang Tan, Liz Jenkins, Louis Martin, Lovish Madaan, Lubo Malo, Lukas Blecher, Lukas Landzaat, Luke de~Oliveira, Madeline Muzzi, Mahesh Pasupuleti, Mannat Singh, Manohar Paluri, Marcin Kardas, Maria Tsimpoukelli, Mathew Oldham, Mathieu Rita, Maya Pavlova, Melanie Kambadur,
  Mike Lewis, Min Si, Mitesh~Kumar Singh, Mona Hassan, Naman Goyal, Narjes Torabi, Nikolay Bashlykov, Nikolay Bogoychev, Niladri Chatterji, Ning Zhang, Olivier Duchenne, Onur Çelebi, Patrick Alrassy, Pengchuan Zhang, Pengwei Li, Petar Vasic, Peter Weng, Prajjwal Bhargava, Pratik Dubal, Praveen Krishnan, Punit~Singh Koura, Puxin Xu, Qing He, Qingxiao Dong, Ragavan Srinivasan, Raj Ganapathy, Ramon Calderer, Ricardo~Silveira Cabral, Robert Stojnic, Roberta Raileanu, Rohan Maheswari, Rohit Girdhar, Rohit Patel, Romain Sauvestre, Ronnie Polidoro, Roshan Sumbaly, Ross Taylor, Ruan Silva, Rui Hou, Rui Wang, Saghar Hosseini, Sahana Chennabasappa, Sanjay Singh, Sean Bell, Seohyun~Sonia Kim, Sergey Edunov, Shaoliang Nie, Sharan Narang, Sharath Raparthy, Sheng Shen, Shengye Wan, Shruti Bhosale, Shun Zhang, Simon Vandenhende, Soumya Batra, Spencer Whitman, Sten Sootla, Stephane Collot, Suchin Gururangan, Sydney Borodinsky, Tamar Herman, Tara Fowler, Tarek Sheasha, Thomas Georgiou, Thomas Scialom, Tobias Speckbacher,
  Todor Mihaylov, Tong Xiao, Ujjwal Karn, Vedanuj Goswami, Vibhor Gupta, Vignesh Ramanathan, Viktor Kerkez, Vincent Gonguet, Virginie Do, Vish Vogeti, Vítor Albiero, Vladan Petrovic, Weiwei Chu, Wenhan Xiong, Wenyin Fu, Whitney Meers, Xavier Martinet, Xiaodong Wang, Xiaofang Wang, Xiaoqing~Ellen Tan, Xide Xia, Xinfeng Xie, Xuchao Jia, Xuewei Wang, Yaelle Goldschlag, Yashesh Gaur, Yasmine Babaei, Yi~Wen, Yiwen Song, Yuchen Zhang, Yue Li, Yuning Mao, Zacharie~Delpierre Coudert, Zheng Yan, Zhengxing Chen, Zoe Papakipos, Aaditya Singh, Aayushi Srivastava, Abha Jain, Adam Kelsey, Adam Shajnfeld, Adithya Gangidi, Adolfo Victoria, Ahuva Goldstand, Ajay Menon, Ajay Sharma, Alex Boesenberg, Alexei Baevski, Allie Feinstein, Amanda Kallet, Amit Sangani, Amos Teo, Anam Yunus, Andrei Lupu, Andres Alvarado, Andrew Caples, Andrew Gu, Andrew Ho, Andrew Poulton, Andrew Ryan, Ankit Ramchandani, Annie Dong, Annie Franco, Anuj Goyal, Aparajita Saraf, Arkabandhu Chowdhury, Ashley Gabriel, Ashwin Bharambe, Assaf Eisenman, Azadeh
  Yazdan, Beau James, Ben Maurer, Benjamin Leonhardi, Bernie Huang, Beth Loyd, Beto~De Paola, Bhargavi Paranjape, Bing Liu, Bo~Wu, Boyu Ni, Braden Hancock, Bram Wasti, Brandon Spence, Brani Stojkovic, Brian Gamido, Britt Montalvo, Carl Parker, Carly Burton, Catalina Mejia, Ce~Liu, Changhan Wang, Changkyu Kim, Chao Zhou, Chester Hu, Ching-Hsiang Chu, Chris Cai, Chris Tindal, Christoph Feichtenhofer, Cynthia Gao, Damon Civin, Dana Beaty, Daniel Kreymer, Daniel Li, David Adkins, David Xu, Davide Testuggine, Delia David, Devi Parikh, Diana Liskovich, Didem Foss, Dingkang Wang, Duc Le, Dustin Holland, Edward Dowling, Eissa Jamil, Elaine Montgomery, Eleonora Presani, Emily Hahn, Emily Wood, Eric-Tuan Le, Erik Brinkman, Esteban Arcaute, Evan Dunbar, Evan Smothers, Fei Sun, Felix Kreuk, Feng Tian, Filippos Kokkinos, Firat Ozgenel, Francesco Caggioni, Frank Kanayet, Frank Seide, Gabriela~Medina Florez, Gabriella Schwarz, Gada Badeer, Georgia Swee, Gil Halpern, Grant Herman, Grigory Sizov, Guangyi, Zhang, Guna
  Lakshminarayanan, Hakan Inan, Hamid Shojanazeri, Han Zou, Hannah Wang, Hanwen Zha, Haroun Habeeb, Harrison Rudolph, Helen Suk, Henry Aspegren, Hunter Goldman, Hongyuan Zhan, Ibrahim Damlaj, Igor Molybog, Igor Tufanov, Ilias Leontiadis, Irina-Elena Veliche, Itai Gat, Jake Weissman, James Geboski, James Kohli, Janice Lam, Japhet Asher, Jean-Baptiste Gaya, Jeff Marcus, Jeff Tang, Jennifer Chan, Jenny Zhen, Jeremy Reizenstein, Jeremy Teboul, Jessica Zhong, Jian Jin, Jingyi Yang, Joe Cummings, Jon Carvill, Jon Shepard, Jonathan McPhie, Jonathan Torres, Josh Ginsburg, Junjie Wang, Kai Wu, Kam~Hou U, Karan Saxena, Kartikay Khandelwal, Katayoun Zand, Kathy Matosich, Kaushik Veeraraghavan, Kelly Michelena, Keqian Li, Kiran Jagadeesh, Kun Huang, Kunal Chawla, Kyle Huang, Lailin Chen, Lakshya Garg, Lavender A, Leandro Silva, Lee Bell, Lei Zhang, Liangpeng Guo, Licheng Yu, Liron Moshkovich, Luca Wehrstedt, Madian Khabsa, Manav Avalani, Manish Bhatt, Martynas Mankus, Matan Hasson, Matthew Lennie, Matthias Reso, Maxim
  Groshev, Maxim Naumov, Maya Lathi, Meghan Keneally, Miao Liu, Michael~L. Seltzer, Michal Valko, Michelle Restrepo, Mihir Patel, Mik Vyatskov, Mikayel Samvelyan, Mike Clark, Mike Macey, Mike Wang, Miquel~Jubert Hermoso, Mo~Metanat, Mohammad Rastegari, Munish Bansal, Nandhini Santhanam, Natascha Parks, Natasha White, Navyata Bawa, Nayan Singhal, Nick Egebo, Nicolas Usunier, Nikhil Mehta, Nikolay~Pavlovich Laptev, Ning Dong, Norman Cheng, Oleg Chernoguz, Olivia Hart, Omkar Salpekar, Ozlem Kalinli, Parkin Kent, Parth Parekh, Paul Saab, Pavan Balaji, Pedro Rittner, Philip Bontrager, Pierre Roux, Piotr Dollar, Polina Zvyagina, Prashant Ratanchandani, Pritish Yuvraj, Qian Liang, Rachad Alao, Rachel Rodriguez, Rafi Ayub, Raghotham Murthy, Raghu Nayani, Rahul Mitra, Rangaprabhu Parthasarathy, Raymond Li, Rebekkah Hogan, Robin Battey, Rocky Wang, Russ Howes, Ruty Rinott, Sachin Mehta, Sachin Siby, Sai~Jayesh Bondu, Samyak Datta, Sara Chugh, Sara Hunt, Sargun Dhillon, Sasha Sidorov, Satadru Pan, Saurabh Mahajan,
  Saurabh Verma, Seiji Yamamoto, Sharadh Ramaswamy, Shaun Lindsay, Shaun Lindsay, Sheng Feng, Shenghao Lin, Shengxin~Cindy Zha, Shishir Patil, Shiva Shankar, Shuqiang Zhang, Shuqiang Zhang, Sinong Wang, Sneha Agarwal, Soji Sajuyigbe, Soumith Chintala, Stephanie Max, Stephen Chen, Steve Kehoe, Steve Satterfield, Sudarshan Govindaprasad, Sumit Gupta, Summer Deng, Sungmin Cho, Sunny Virk, Suraj Subramanian, Sy~Choudhury, Sydney Goldman, Tal Remez, Tamar Glaser, Tamara Best, Thilo Koehler, Thomas Robinson, Tianhe Li, Tianjun Zhang, Tim Matthews, Timothy Chou, Tzook Shaked, Varun Vontimitta, Victoria Ajayi, Victoria Montanez, Vijai Mohan, Vinay~Satish Kumar, Vishal Mangla, Vlad Ionescu, Vlad Poenaru, Vlad~Tiberiu Mihailescu, Vladimir Ivanov, Wei Li, Wenchen Wang, Wenwen Jiang, Wes Bouaziz, Will Constable, Xiaocheng Tang, Xiaojian Wu, Xiaolan Wang, Xilun Wu, Xinbo Gao, Yaniv Kleinman, Yanjun Chen, Ye~Hu, Ye~Jia, Ye~Qi, Yenda Li, Yilin Zhang, Ying Zhang, Yossi Adi, Youngjin Nam, Yu, Wang, Yu~Zhao, Yuchen Hao, Yundi
  Qian, Yunlu Li, Yuzi He, Zach Rait, Zachary DeVito, Zef Rosnbrick, Zhaoduo Wen, Zhenyu Yang, Zhiwei Zhao, and Zhiyu Ma. 2024.
\newblock \href {https://arxiv.org/abs/2407.21783} {The llama 3 herd of models}.
\newblock \emph{Preprint}, arXiv:2407.21783.

\bibitem[{Gudibande et~al.(2023)Gudibande, Wallace, Snell, Geng, Liu, Abbeel, Levine, and Song}]{gudibande2023false}
Arnav Gudibande, Eric Wallace, Charlie Snell, Xinyang Geng, Hao Liu, Pieter Abbeel, Sergey Levine, and Dawn Song. 2023.
\newblock The false promise of imitating proprietary llms.
\newblock \emph{arXiv preprint arXiv:2305.15717}.

\bibitem[{Guo et~al.(2024)Guo, Shang, Vazirgiannis, and Clavel}]{guo-etal-2024-curious}
Yanzhu Guo, Guokan Shang, Michalis Vazirgiannis, and Chlo{\'e} Clavel. 2024.
\newblock \href {https://doi.org/10.18653/v1/2024.findings-naacl.228} {The curious decline of linguistic diversity: Training language models on synthetic text}.
\newblock In \emph{Findings of the Association for Computational Linguistics: NAACL 2024}, pages 3589--3604, Mexico City, Mexico. Association for Computational Linguistics.

\bibitem[{Jiang et~al.(2023)Jiang, Sablayrolles, Mensch, Bamford, Chaplot, de~las Casas, Bressand, Lengyel, Lample, Saulnier, Lavaud, Lachaux, Stock, Scao, Lavril, Wang, Lacroix, and Sayed}]{jiang2023mistral7b}
Albert~Q. Jiang, Alexandre Sablayrolles, Arthur Mensch, Chris Bamford, Devendra~Singh Chaplot, Diego de~las Casas, Florian Bressand, Gianna Lengyel, Guillaume Lample, Lucile Saulnier, Lélio~Renard Lavaud, Marie-Anne Lachaux, Pierre Stock, Teven~Le Scao, Thibaut Lavril, Thomas Wang, Timothée Lacroix, and William~El Sayed. 2023.
\newblock \href {https://arxiv.org/abs/2310.06825} {Mistral 7b}.
\newblock \emph{Preprint}, arXiv:2310.06825.

\bibitem[{Jin et~al.(2023)Jin, Mehri, Hazarika, Padmakumar, LEE, Liu, and Namazifar}]{jin2023dataefficient}
Di~Jin, Shikib Mehri, Devamanyu Hazarika, Aishwarya Padmakumar, SUNGJIN LEE, Yang Liu, and Mahdi Namazifar. 2023.
\newblock \href {https://openreview.net/forum?id=IPJqprsrNX} {Data-efficient alignment of large language models with human feedback through natural language}.
\newblock In \emph{NeurIPS 2023 Workshop on Instruction Tuning and Instruction Following}.

\bibitem[{Kaur et~al.(2024)Kaur, Park, Goyal, and Arora}]{kaur2024instruct}
Simran Kaur, Simon Park, Anirudh Goyal, and Sanjeev Arora. 2024.
\newblock Instruct-skillmix: A powerful pipeline for llm instruction tuning.
\newblock \emph{arXiv preprint arXiv:2408.14774}.

\bibitem[{Li et~al.(2024{\natexlab{a}})Li, Dong, Tang, Wang, Zhang, Huang, Huang, Huang, Huang, Zhang, Gu, Cheng, Wang, Chen, Dong, Lu, Sui, Wang, Lam, and Wei}]{li2024syntheticdataalmostscratch}
Haoran Li, Qingxiu Dong, Zhengyang Tang, Chaojun Wang, Xingxing Zhang, Haoyang Huang, Shaohan Huang, Xiaolong Huang, Zeqiang Huang, Dongdong Zhang, Yuxian Gu, Xin Cheng, Xun Wang, Si-Qing Chen, Li~Dong, Wei Lu, Zhifang Sui, Benyou Wang, Wai Lam, and Furu Wei. 2024{\natexlab{a}}.
\newblock \href {https://arxiv.org/abs/2402.13064} {Synthetic data (almost) from scratch: Generalized instruction tuning for language models}.
\newblock \emph{Preprint}, arXiv:2402.13064.

\bibitem[{Li et~al.(2025)Li, Du, Zhao, wen Zhang, Wang, Gao, Liu, and Lin}]{InfinityInstruct2024}
Jijie Li, Li~Du, Hanyu Zhao, Bo~wen Zhang, Liangdong Wang, Boyan Gao, Guang Liu, and Yonghua Lin. 2025.
\newblock \href {https://arxiv.org/abs/2506.11116} {Infinity instruct: Scaling instruction selection and synthesis to enhance language models}.
\newblock \emph{Preprint}, arXiv:2506.11116.

\bibitem[{Li et~al.(2024{\natexlab{b}})Li, Chiang, Frick, Dunlap, Wu, Zhu, Gonzalez, and Stoica}]{li2024crowdsourced}
Tianle Li, Wei-Lin Chiang, Evan Frick, Lisa Dunlap, Tianhao Wu, Banghua Zhu, Joseph~E. Gonzalez, and Ion Stoica. 2024{\natexlab{b}}.
\newblock \href {https://arxiv.org/abs/2406.11939} {From crowdsourced data to high-quality benchmarks: Arena-hard and benchbuilder pipeline}.
\newblock \emph{Preprint}, arXiv:2406.11939.

\bibitem[{Li et~al.(2024{\natexlab{c}})Li, Chiang, Frick, Dunlap, Zhu, Gonzalez, and Stoica}]{arenahard2024}
Tianle Li, Wei-Lin Chiang, Evan Frick, Lisa Dunlap, Banghua Zhu, Joseph~E. Gonzalez, and Ion Stoica. 2024{\natexlab{c}}.
\newblock \href {https://lmsys.org/blog/2024-04-19-arena-hard/} {From live data to high-quality benchmarks: The arena-hard pipeline}.

\bibitem[{Madaan et~al.(2023)Madaan, Tandon, Gupta, Hallinan, Gao, Wiegreffe, Alon, Dziri, Prabhumoye, Yang, Gupta, Majumder, Hermann, Welleck, Yazdanbakhsh, and Clark}]{NEURIPS2023_91edff07}
Aman Madaan, Niket Tandon, Prakhar Gupta, Skyler Hallinan, Luyu Gao, Sarah Wiegreffe, Uri Alon, Nouha Dziri, Shrimai Prabhumoye, Yiming Yang, Shashank Gupta, Bodhisattwa~Prasad Majumder, Katherine Hermann, Sean Welleck, Amir Yazdanbakhsh, and Peter Clark. 2023.
\newblock \href {https://proceedings.neurips.cc/paper_files/paper/2023/file/91edff07232fb1b55a505a9e9f6c0ff3-Paper-Conference.pdf} {Self-refine: Iterative refinement with self-feedback}.
\newblock In \emph{Advances in Neural Information Processing Systems}, volume~36, pages 46534--46594. Curran Associates, Inc.

\bibitem[{Mehri and Shwartz(2023)}]{mehri2023automatic}
Shuhaib Mehri and Vered Shwartz. 2023.
\newblock Automatic evaluation of generative models with instruction tuning.
\newblock In \emph{Proceedings of the Third Workshop on Natural Language Generation, Evaluation, and Metrics (GEM)}, pages 42--52.

\bibitem[{OpenAI et~al.(2024{\natexlab{a}})OpenAI, :, Hurst, Lerer, Goucher, Perelman, Ramesh, Clark, Ostrow, Welihinda, Hayes, Radford, Mądry, Baker-Whitcomb, Beutel, Borzunov, Carney, Chow, Kirillov, Nichol, Paino, Renzin, Passos, Kirillov, Christakis, Conneau, Kamali, Jabri, Moyer, Tam, Crookes, Tootoochian, Tootoonchian, Kumar, Vallone, Karpathy, Braunstein, Cann, Codispoti, Galu, Kondrich, Tulloch, Mishchenko, Baek, Jiang, Pelisse, Woodford, Gosalia, Dhar, Pantuliano, Nayak, Oliver, Zoph, Ghorbani, Leimberger, Rossen, Sokolowsky, Wang, Zweig, Hoover, Samic, McGrew, Spero, Giertler, Cheng, Lightcap, Walkin, Quinn, Guarraci, Hsu, Kellogg, Eastman, Lugaresi, Wainwright, Bassin, Hudson, Chu, Nelson, Li, Shern, Conger, Barette, Voss, Ding, Lu, Zhang, Beaumont, Hallacy, Koch, Gibson, Kim, Choi, McLeavey, Hesse, Fischer, Winter, Czarnecki, Jarvis, Wei, Koumouzelis, Sherburn, Kappler, Levin, Levy, Carr, Farhi, Mely, Robinson, Sasaki, Jin, Valladares, Tsipras, Li, Nguyen, Findlay, Oiwoh, Wong, Asdar, Proehl,
  Yang, Antonow, Kramer, Peterson, Sigler, Wallace, Brevdo, Mays, Khorasani, Such, Raso, Zhang, von Lohmann, Sulit, Goh, Oden, Salmon, Starace, Brockman, Salman, Bao, Hu, Wong, Wang, Schmidt, Whitney, Jun, Kirchner, de~Oliveira~Pinto, Ren, Chang, Chung, Kivlichan, O'Connell, O'Connell, Osband, Silber, Sohl, Okuyucu, Lan, Kostrikov, Sutskever, Kanitscheider, Gulrajani, Coxon, Menick, Pachocki, Aung, Betker, Crooks, Lennon, Kiros, Leike, Park, Kwon, Phang, Teplitz, Wei, Wolfe, Chen, Harris, Varavva, Lee, Shieh, Lin, Yu, Weng, Tang, Yu, Jang, Candela, Beutler, Landers, Parish, Heidecke, Schulman, Lachman, McKay, Uesato, Ward, Kim, Huizinga, Sitkin, Kraaijeveld, Gross, Kaplan, Snyder, Achiam, Jiao, Lee, Zhuang, Harriman, Fricke, Hayashi, Singhal, Shi, Karthik, Wood, Rimbach, Hsu, Nguyen, Gu-Lemberg, Button, Liu, Howe, Muthukumar, Luther, Ahmad, Kai, Itow, Workman, Pathak, Chen, Jing, Guy, Fedus, Zhou, Mamitsuka, Weng, McCallum, Held, Ouyang, Feuvrier, Zhang, Kondraciuk, Kaiser, Hewitt, Metz, Doshi, Aflak, Simens,
  Boyd, Thompson, Dukhan, Chen, Gray, Hudnall, Zhang, Aljubeh, Litwin, Zeng, Johnson, Shetty, Gupta, Shah, Yatbaz, Yang, Zhong, Glaese, Chen, Janner, Lampe, Petrov, Wu, Wang, Fradin, Pokrass, Castro, de~Castro, Pavlov, Brundage, Wang, Khan, Murati, Bavarian, Lin, Yesildal, Soto, Gimelshein, Cone, Staudacher, Summers, LaFontaine, Chowdhury, Ryder, Stathas, Turley, Tezak, Felix, Kudige, Keskar, Deutsch, Bundick, Puckett, Nachum, Okelola, Boiko, Murk, Jaffe, Watkins, Godement, Campbell-Moore, Chao, McMillan, Belov, Su, Bak, Bakkum, Deng, Dolan, Hoeschele, Welinder, Tillet, Pronin, Tillet, Dhariwal, Yuan, Dias, Lim, Arora, Troll, Lin, Lopes, Puri, Miyara, Leike, Gaubert, Zamani, Wang, Donnelly, Honsby, Smith, Sahai, Ramchandani, Huet, Carmichael, Zellers, Chen, Chen, Nigmatullin, Cheu, Jain, Altman, Schoenholz, Toizer, Miserendino, Agarwal, Culver, Ethersmith, Gray, Grove, Metzger, Hermani, Jain, Zhao, Wu, Jomoto, Wu, Shuaiqi, Xia, Phene, Papay, Narayanan, Coffey, Lee, Hall, Balaji, Broda, Stramer, Xu, Gogineni,
  Christianson, Sanders, Patwardhan, Cunninghman, Degry, Dimson, Raoux, Shadwell, Zheng, Underwood, Markov, Sherbakov, Rubin, Stasi, Kaftan, Heywood, Peterson, Walters, Eloundou, Qi, Moeller, Monaco, Kuo, Fomenko, Chang, Zheng, Zhou, Manassra, Sheu, Zaremba, Patil, Qian, Kim, Cheng, Zhang, He, Zhang, Jin, Dai, and Malkov}]{openai2024gpt4ocard}
OpenAI, :, Aaron Hurst, Adam Lerer, Adam~P. Goucher, Adam Perelman, Aditya Ramesh, Aidan Clark, AJ~Ostrow, Akila Welihinda, Alan Hayes, Alec Radford, Aleksander Mądry, Alex Baker-Whitcomb, Alex Beutel, Alex Borzunov, Alex Carney, Alex Chow, Alex Kirillov, Alex Nichol, Alex Paino, Alex Renzin, Alex~Tachard Passos, Alexander Kirillov, Alexi Christakis, Alexis Conneau, Ali Kamali, Allan Jabri, Allison Moyer, Allison Tam, Amadou Crookes, Amin Tootoochian, Amin Tootoonchian, Ananya Kumar, Andrea Vallone, Andrej Karpathy, Andrew Braunstein, Andrew Cann, Andrew Codispoti, Andrew Galu, Andrew Kondrich, Andrew Tulloch, Andrey Mishchenko, Angela Baek, Angela Jiang, Antoine Pelisse, Antonia Woodford, Anuj Gosalia, Arka Dhar, Ashley Pantuliano, Avi Nayak, Avital Oliver, Barret Zoph, Behrooz Ghorbani, Ben Leimberger, Ben Rossen, Ben Sokolowsky, Ben Wang, Benjamin Zweig, Beth Hoover, Blake Samic, Bob McGrew, Bobby Spero, Bogo Giertler, Bowen Cheng, Brad Lightcap, Brandon Walkin, Brendan Quinn, Brian Guarraci, Brian Hsu,
  Bright Kellogg, Brydon Eastman, Camillo Lugaresi, Carroll Wainwright, Cary Bassin, Cary Hudson, Casey Chu, Chad Nelson, Chak Li, Chan~Jun Shern, Channing Conger, Charlotte Barette, Chelsea Voss, Chen Ding, Cheng Lu, Chong Zhang, Chris Beaumont, Chris Hallacy, Chris Koch, Christian Gibson, Christina Kim, Christine Choi, Christine McLeavey, Christopher Hesse, Claudia Fischer, Clemens Winter, Coley Czarnecki, Colin Jarvis, Colin Wei, Constantin Koumouzelis, Dane Sherburn, Daniel Kappler, Daniel Levin, Daniel Levy, David Carr, David Farhi, David Mely, David Robinson, David Sasaki, Denny Jin, Dev Valladares, Dimitris Tsipras, Doug Li, Duc~Phong Nguyen, Duncan Findlay, Edede Oiwoh, Edmund Wong, Ehsan Asdar, Elizabeth Proehl, Elizabeth Yang, Eric Antonow, Eric Kramer, Eric Peterson, Eric Sigler, Eric Wallace, Eugene Brevdo, Evan Mays, Farzad Khorasani, Felipe~Petroski Such, Filippo Raso, Francis Zhang, Fred von Lohmann, Freddie Sulit, Gabriel Goh, Gene Oden, Geoff Salmon, Giulio Starace, Greg Brockman, Hadi
  Salman, Haiming Bao, Haitang Hu, Hannah Wong, Haoyu Wang, Heather Schmidt, Heather Whitney, Heewoo Jun, Hendrik Kirchner, Henrique~Ponde de~Oliveira~Pinto, Hongyu Ren, Huiwen Chang, Hyung~Won Chung, Ian Kivlichan, Ian O'Connell, Ian O'Connell, Ian Osband, Ian Silber, Ian Sohl, Ibrahim Okuyucu, Ikai Lan, Ilya Kostrikov, Ilya Sutskever, Ingmar Kanitscheider, Ishaan Gulrajani, Jacob Coxon, Jacob Menick, Jakub Pachocki, James Aung, James Betker, James Crooks, James Lennon, Jamie Kiros, Jan Leike, Jane Park, Jason Kwon, Jason Phang, Jason Teplitz, Jason Wei, Jason Wolfe, Jay Chen, Jeff Harris, Jenia Varavva, Jessica~Gan Lee, Jessica Shieh, Ji~Lin, Jiahui Yu, Jiayi Weng, Jie Tang, Jieqi Yu, Joanne Jang, Joaquin~Quinonero Candela, Joe Beutler, Joe Landers, Joel Parish, Johannes Heidecke, John Schulman, Jonathan Lachman, Jonathan McKay, Jonathan Uesato, Jonathan Ward, Jong~Wook Kim, Joost Huizinga, Jordan Sitkin, Jos Kraaijeveld, Josh Gross, Josh Kaplan, Josh Snyder, Joshua Achiam, Joy Jiao, Joyce Lee, Juntang
  Zhuang, Justyn Harriman, Kai Fricke, Kai Hayashi, Karan Singhal, Katy Shi, Kavin Karthik, Kayla Wood, Kendra Rimbach, Kenny Hsu, Kenny Nguyen, Keren Gu-Lemberg, Kevin Button, Kevin Liu, Kiel Howe, Krithika Muthukumar, Kyle Luther, Lama Ahmad, Larry Kai, Lauren Itow, Lauren Workman, Leher Pathak, Leo Chen, Li~Jing, Lia Guy, Liam Fedus, Liang Zhou, Lien Mamitsuka, Lilian Weng, Lindsay McCallum, Lindsey Held, Long Ouyang, Louis Feuvrier, Lu~Zhang, Lukas Kondraciuk, Lukasz Kaiser, Luke Hewitt, Luke Metz, Lyric Doshi, Mada Aflak, Maddie Simens, Madelaine Boyd, Madeleine Thompson, Marat Dukhan, Mark Chen, Mark Gray, Mark Hudnall, Marvin Zhang, Marwan Aljubeh, Mateusz Litwin, Matthew Zeng, Max Johnson, Maya Shetty, Mayank Gupta, Meghan Shah, Mehmet Yatbaz, Meng~Jia Yang, Mengchao Zhong, Mia Glaese, Mianna Chen, Michael Janner, Michael Lampe, Michael Petrov, Michael Wu, Michele Wang, Michelle Fradin, Michelle Pokrass, Miguel Castro, Miguel Oom~Temudo de~Castro, Mikhail Pavlov, Miles Brundage, Miles Wang, Minal
  Khan, Mira Murati, Mo~Bavarian, Molly Lin, Murat Yesildal, Nacho Soto, Natalia Gimelshein, Natalie Cone, Natalie Staudacher, Natalie Summers, Natan LaFontaine, Neil Chowdhury, Nick Ryder, Nick Stathas, Nick Turley, Nik Tezak, Niko Felix, Nithanth Kudige, Nitish Keskar, Noah Deutsch, Noel Bundick, Nora Puckett, Ofir Nachum, Ola Okelola, Oleg Boiko, Oleg Murk, Oliver Jaffe, Olivia Watkins, Olivier Godement, Owen Campbell-Moore, Patrick Chao, Paul McMillan, Pavel Belov, Peng Su, Peter Bak, Peter Bakkum, Peter Deng, Peter Dolan, Peter Hoeschele, Peter Welinder, Phil Tillet, Philip Pronin, Philippe Tillet, Prafulla Dhariwal, Qiming Yuan, Rachel Dias, Rachel Lim, Rahul Arora, Rajan Troll, Randall Lin, Rapha~Gontijo Lopes, Raul Puri, Reah Miyara, Reimar Leike, Renaud Gaubert, Reza Zamani, Ricky Wang, Rob Donnelly, Rob Honsby, Rocky Smith, Rohan Sahai, Rohit Ramchandani, Romain Huet, Rory Carmichael, Rowan Zellers, Roy Chen, Ruby Chen, Ruslan Nigmatullin, Ryan Cheu, Saachi Jain, Sam Altman, Sam Schoenholz, Sam
  Toizer, Samuel Miserendino, Sandhini Agarwal, Sara Culver, Scott Ethersmith, Scott Gray, Sean Grove, Sean Metzger, Shamez Hermani, Shantanu Jain, Shengjia Zhao, Sherwin Wu, Shino Jomoto, Shirong Wu, Shuaiqi, Xia, Sonia Phene, Spencer Papay, Srinivas Narayanan, Steve Coffey, Steve Lee, Stewart Hall, Suchir Balaji, Tal Broda, Tal Stramer, Tao Xu, Tarun Gogineni, Taya Christianson, Ted Sanders, Tejal Patwardhan, Thomas Cunninghman, Thomas Degry, Thomas Dimson, Thomas Raoux, Thomas Shadwell, Tianhao Zheng, Todd Underwood, Todor Markov, Toki Sherbakov, Tom Rubin, Tom Stasi, Tomer Kaftan, Tristan Heywood, Troy Peterson, Tyce Walters, Tyna Eloundou, Valerie Qi, Veit Moeller, Vinnie Monaco, Vishal Kuo, Vlad Fomenko, Wayne Chang, Weiyi Zheng, Wenda Zhou, Wesam Manassra, Will Sheu, Wojciech Zaremba, Yash Patil, Yilei Qian, Yongjik Kim, Youlong Cheng, Yu~Zhang, Yuchen He, Yuchen Zhang, Yujia Jin, Yunxing Dai, and Yury Malkov. 2024{\natexlab{a}}.
\newblock \href {https://arxiv.org/abs/2410.21276} {Gpt-4o system card}.
\newblock \emph{Preprint}, arXiv:2410.21276.

\bibitem[{OpenAI et~al.(2024{\natexlab{b}})OpenAI, :, Jaech, Kalai, Lerer, Richardson, El-Kishky, Low, Helyar, Madry, Beutel, Carney, Iftimie, Karpenko, Passos, Neitz, Prokofiev, Wei, Tam, Bennett, Kumar, Saraiva, Vallone, Duberstein, Kondrich, Mishchenko, Applebaum, Jiang, Nair, Zoph, Ghorbani, Rossen, Sokolowsky, Barak, McGrew, Minaiev, Hao, Baker, Houghton, McKinzie, Eastman, Lugaresi, Bassin, Hudson, Li, de~Bourcy, Voss, Shen, Zhang, Koch, Orsinger, Hesse, Fischer, Chan, Roberts, Kappler, Levy, Selsam, Dohan, Farhi, Mely, Robinson, Tsipras, Li, Oprica, Freeman, Zhang, Wong, Proehl, Cheung, Mitchell, Wallace, Ritter, Mays, Wang, Such, Raso, Leoni, Tsimpourlas, Song, von Lohmann, Sulit, Salmon, Parascandolo, Chabot, Zhao, Brockman, Leclerc, Salman, Bao, Sheng, Andrin, Bagherinezhad, Ren, Lightman, Chung, Kivlichan, O'Connell, Osband, Gilaberte, Akkaya, Kostrikov, Sutskever, Kofman, Pachocki, Lennon, Wei, Harb, Twore, Feng, Yu, Weng, Tang, Yu, Candela, Palermo, Parish, Heidecke, Hallman, Rizzo, Gordon,
  Uesato, Ward, Huizinga, Wang, Chen, Xiao, Singhal, Nguyen, Cobbe, Shi, Wood, Rimbach, Gu-Lemberg, Liu, Lu, Stone, Yu, Ahmad, Yang, Liu, Maksin, Ho, Fedus, Weng, Li, McCallum, Held, Kuhn, Kondraciuk, Kaiser, Metz, Boyd, Trebacz, Joglekar, Chen, Tintor, Meyer, Jones, Kaufer, Schwarzer, Shah, Yatbaz, Guan, Xu, Yan, Glaese, Chen, Lampe, Malek, Wang, Fradin, McClay, Pavlov, Wang, Wang, Murati, Bavarian, Rohaninejad, McAleese, Chowdhury, Chowdhury, Ryder, Tezak, Brown, Nachum, Boiko, Murk, Watkins, Chao, Ashbourne, Izmailov, Zhokhov, Dias, Arora, Lin, Lopes, Gaon, Miyara, Leike, Hwang, Garg, Brown, James, Shu, Cheu, Greene, Jain, Altman, Toizer, Toyer, Miserendino, Agarwal, Hernandez, Baker, McKinney, Yan, Zhao, Hu, Santurkar, Chaudhuri, Zhang, Fu, Papay, Lin, Balaji, Sanjeev, Sidor, Broda, Clark, Wang, Gordon, Sanders, Patwardhan, Sottiaux, Degry, Dimson, Zheng, Garipov, Stasi, Bansal, Creech, Peterson, Eloundou, Qi, Kosaraju, Monaco, Pong, Fomenko, Zheng, Zhou, McCabe, Zaremba, Dubois, Lu, Chen, Cha, Bai, He,
  Zhang, Wang, Shao, and Li}]{openai2024openaio1card}
OpenAI, :, Aaron Jaech, Adam Kalai, Adam Lerer, Adam Richardson, Ahmed El-Kishky, Aiden Low, Alec Helyar, Aleksander Madry, Alex Beutel, Alex Carney, Alex Iftimie, Alex Karpenko, Alex~Tachard Passos, Alexander Neitz, Alexander Prokofiev, Alexander Wei, Allison Tam, Ally Bennett, Ananya Kumar, Andre Saraiva, Andrea Vallone, Andrew Duberstein, Andrew Kondrich, Andrey Mishchenko, Andy Applebaum, Angela Jiang, Ashvin Nair, Barret Zoph, Behrooz Ghorbani, Ben Rossen, Benjamin Sokolowsky, Boaz Barak, Bob McGrew, Borys Minaiev, Botao Hao, Bowen Baker, Brandon Houghton, Brandon McKinzie, Brydon Eastman, Camillo Lugaresi, Cary Bassin, Cary Hudson, Chak~Ming Li, Charles de~Bourcy, Chelsea Voss, Chen Shen, Chong Zhang, Chris Koch, Chris Orsinger, Christopher Hesse, Claudia Fischer, Clive Chan, Dan Roberts, Daniel Kappler, Daniel Levy, Daniel Selsam, David Dohan, David Farhi, David Mely, David Robinson, Dimitris Tsipras, Doug Li, Dragos Oprica, Eben Freeman, Eddie Zhang, Edmund Wong, Elizabeth Proehl, Enoch Cheung, Eric
  Mitchell, Eric Wallace, Erik Ritter, Evan Mays, Fan Wang, Felipe~Petroski Such, Filippo Raso, Florencia Leoni, Foivos Tsimpourlas, Francis Song, Fred von Lohmann, Freddie Sulit, Geoff Salmon, Giambattista Parascandolo, Gildas Chabot, Grace Zhao, Greg Brockman, Guillaume Leclerc, Hadi Salman, Haiming Bao, Hao Sheng, Hart Andrin, Hessam Bagherinezhad, Hongyu Ren, Hunter Lightman, Hyung~Won Chung, Ian Kivlichan, Ian O'Connell, Ian Osband, Ignasi~Clavera Gilaberte, Ilge Akkaya, Ilya Kostrikov, Ilya Sutskever, Irina Kofman, Jakub Pachocki, James Lennon, Jason Wei, Jean Harb, Jerry Twore, Jiacheng Feng, Jiahui Yu, Jiayi Weng, Jie Tang, Jieqi Yu, Joaquin~Quiñonero Candela, Joe Palermo, Joel Parish, Johannes Heidecke, John Hallman, John Rizzo, Jonathan Gordon, Jonathan Uesato, Jonathan Ward, Joost Huizinga, Julie Wang, Kai Chen, Kai Xiao, Karan Singhal, Karina Nguyen, Karl Cobbe, Katy Shi, Kayla Wood, Kendra Rimbach, Keren Gu-Lemberg, Kevin Liu, Kevin Lu, Kevin Stone, Kevin Yu, Lama Ahmad, Lauren Yang, Leo Liu,
  Leon Maksin, Leyton Ho, Liam Fedus, Lilian Weng, Linden Li, Lindsay McCallum, Lindsey Held, Lorenz Kuhn, Lukas Kondraciuk, Lukasz Kaiser, Luke Metz, Madelaine Boyd, Maja Trebacz, Manas Joglekar, Mark Chen, Marko Tintor, Mason Meyer, Matt Jones, Matt Kaufer, Max Schwarzer, Meghan Shah, Mehmet Yatbaz, Melody~Y. Guan, Mengyuan Xu, Mengyuan Yan, Mia Glaese, Mianna Chen, Michael Lampe, Michael Malek, Michele Wang, Michelle Fradin, Mike McClay, Mikhail Pavlov, Miles Wang, Mingxuan Wang, Mira Murati, Mo~Bavarian, Mostafa Rohaninejad, Nat McAleese, Neil Chowdhury, Neil Chowdhury, Nick Ryder, Nikolas Tezak, Noam Brown, Ofir Nachum, Oleg Boiko, Oleg Murk, Olivia Watkins, Patrick Chao, Paul Ashbourne, Pavel Izmailov, Peter Zhokhov, Rachel Dias, Rahul Arora, Randall Lin, Rapha~Gontijo Lopes, Raz Gaon, Reah Miyara, Reimar Leike, Renny Hwang, Rhythm Garg, Robin Brown, Roshan James, Rui Shu, Ryan Cheu, Ryan Greene, Saachi Jain, Sam Altman, Sam Toizer, Sam Toyer, Samuel Miserendino, Sandhini Agarwal, Santiago Hernandez,
  Sasha Baker, Scott McKinney, Scottie Yan, Shengjia Zhao, Shengli Hu, Shibani Santurkar, Shraman~Ray Chaudhuri, Shuyuan Zhang, Siyuan Fu, Spencer Papay, Steph Lin, Suchir Balaji, Suvansh Sanjeev, Szymon Sidor, Tal Broda, Aidan Clark, Tao Wang, Taylor Gordon, Ted Sanders, Tejal Patwardhan, Thibault Sottiaux, Thomas Degry, Thomas Dimson, Tianhao Zheng, Timur Garipov, Tom Stasi, Trapit Bansal, Trevor Creech, Troy Peterson, Tyna Eloundou, Valerie Qi, Vineet Kosaraju, Vinnie Monaco, Vitchyr Pong, Vlad Fomenko, Weiyi Zheng, Wenda Zhou, Wes McCabe, Wojciech Zaremba, Yann Dubois, Yinghai Lu, Yining Chen, Young Cha, Yu~Bai, Yuchen He, Yuchen Zhang, Yunyun Wang, Zheng Shao, and Zhuohan Li. 2024{\natexlab{b}}.
\newblock \href {https://arxiv.org/abs/2412.16720} {Openai o1 system card}.
\newblock \emph{Preprint}, arXiv:2412.16720.

\bibitem[{OpenAI(2024)}]{OpenAIGpt4oMini}
OpenAI. 2024.
\newblock \href {https://openai.com/index/gpt-4o-mini-advancing-cost-efficient-intelligence/} {Gpt-4o mini: advancing cost-efficient intelligence}.

\bibitem[{Ouyang et~al.(2022)Ouyang, Wu, Jiang, Almeida, Wainwright, Mishkin, Zhang, Agarwal, Slama, Ray, Schulman, Hilton, Kelton, Miller, Simens, Askell, Welinder, Christiano, Leike, and Lowe}]{NEURIPS2022_b1efde53}
Long Ouyang, Jeffrey Wu, Xu~Jiang, Diogo Almeida, Carroll Wainwright, Pamela Mishkin, Chong Zhang, Sandhini Agarwal, Katarina Slama, Alex Ray, John Schulman, Jacob Hilton, Fraser Kelton, Luke Miller, Maddie Simens, Amanda Askell, Peter Welinder, Paul~F Christiano, Jan Leike, and Ryan Lowe. 2022.
\newblock \href {https://proceedings.neurips.cc/paper_files/paper/2022/file/b1efde53be364a73914f58805a001731-Paper-Conference.pdf} {Training language models to follow instructions with human feedback}.
\newblock In \emph{Advances in Neural Information Processing Systems}, volume~35, pages 27730--27744. Curran Associates, Inc.

\bibitem[{Peng et~al.(2023)Peng, Li, He, Galley, and Gao}]{peng2023instructiontuninggpt4}
Baolin Peng, Chunyuan Li, Pengcheng He, Michel Galley, and Jianfeng Gao. 2023.
\newblock \href {https://arxiv.org/abs/2304.03277} {Instruction tuning with gpt-4}.
\newblock \emph{Preprint}, arXiv:2304.03277.

\bibitem[{Sun et~al.(2023)Sun, Shen, Zhou, Zhang, Chen, Cox, Yang, and Gan}]{NEURIPS2023_0764db11}
Zhiqing Sun, Yikang Shen, Qinhong Zhou, Hongxin Zhang, Zhenfang Chen, David Cox, Yiming Yang, and Chuang Gan. 2023.
\newblock \href {https://proceedings.neurips.cc/paper_files/paper/2023/file/0764db1151b936aca59249e2c1386101-Paper-Conference.pdf} {Principle-driven self-alignment of language models from scratch with minimal human supervision}.
\newblock In \emph{Advances in Neural Information Processing Systems}, volume~36, pages 2511--2565. Curran Associates, Inc.

\bibitem[{Sun et~al.(2024)Sun, Shen, Zhou, Zhang, Chen, Cox, Yang, and Gan}]{sun2024principle}
Zhiqing Sun, Yikang Shen, Qinhong Zhou, Hongxin Zhang, Zhenfang Chen, David Cox, Yiming Yang, and Chuang Gan. 2024.
\newblock Principle-driven self-alignment of language models from scratch with minimal human supervision.
\newblock \emph{Advances in Neural Information Processing Systems}, 36.

\bibitem[{Tang et~al.(2024)Tang, Pang, Liu, Tang, Ye, Dong, Wang, and Chen}]{tang2024synthesizingposttrainingdatallms}
Shuo Tang, Xianghe Pang, Zexi Liu, Bohan Tang, Rui Ye, Xiaowen Dong, Yanfeng Wang, and Siheng Chen. 2024.
\newblock \href {https://arxiv.org/abs/2410.14251} {Synthesizing post-training data for llms through multi-agent simulation}.
\newblock \emph{Preprint}, arXiv:2410.14251.

\bibitem[{Taori et~al.(2023)Taori, Gulrajani, Zhang, Dubois, Li, Guestrin, Liang, and Hashimoto}]{alpaca}
Rohan Taori, Ishaan Gulrajani, Tianyi Zhang, Yann Dubois, Xuechen Li, Carlos Guestrin, Percy Liang, and Tatsunori~B. Hashimoto. 2023.
\newblock Stanford alpaca: An instruction-following llama model.
\newblock \url{https://github.com/tatsu-lab/stanford_alpaca}.

\bibitem[{Wang et~al.(2023{\natexlab{a}})Wang, Yu, Tan, O'Brien, Pasunuru, Dwivedi-Yu, Golovneva, Zettlemoyer, Fazel-Zarandi, and Celikyilmaz}]{wang2023shepherdcriticlanguagemodel}
Tianlu Wang, Ping Yu, Xiaoqing~Ellen Tan, Sean O'Brien, Ramakanth Pasunuru, Jane Dwivedi-Yu, Olga Golovneva, Luke Zettlemoyer, Maryam Fazel-Zarandi, and Asli Celikyilmaz. 2023{\natexlab{a}}.
\newblock \href {https://arxiv.org/abs/2308.04592} {Shepherd: A critic for language model generation}.
\newblock \emph{Preprint}, arXiv:2308.04592.

\bibitem[{Wang et~al.(2023{\natexlab{b}})Wang, Kordi, Mishra, Liu, Smith, Khashabi, and Hajishirzi}]{wang-etal-2023-self-instruct}
Yizhong Wang, Yeganeh Kordi, Swaroop Mishra, Alisa Liu, Noah~A. Smith, Daniel Khashabi, and Hannaneh Hajishirzi. 2023{\natexlab{b}}.
\newblock \href {https://doi.org/10.18653/v1/2023.acl-long.754} {Self-instruct: Aligning language models with self-generated instructions}.
\newblock pages 13484--13508, Toronto, Canada.

\bibitem[{Wu et~al.(2024)Wu, Yuan, Golovneva, Xu, Tian, Jiao, Weston, and Sukhbaatar}]{wu2024meta}
Tianhao Wu, Weizhe Yuan, Olga Golovneva, Jing Xu, Yuandong Tian, Jiantao Jiao, Jason Weston, and Sainbayar Sukhbaatar. 2024.
\newblock Meta-rewarding language models: Self-improving alignment with llm-as-a-meta-judge.
\newblock \emph{arXiv preprint arXiv:2407.19594}.

\bibitem[{Xu et~al.(2023)Xu, Sun, Zheng, Geng, Zhao, Feng, Tao, and Jiang}]{xu2023wizardlmempoweringlargelanguage}
Can Xu, Qingfeng Sun, Kai Zheng, Xiubo Geng, Pu~Zhao, Jiazhan Feng, Chongyang Tao, and Daxin Jiang. 2023.
\newblock \href {https://arxiv.org/abs/2304.12244} {Wizardlm: Empowering large language models to follow complex instructions}.
\newblock \emph{Preprint}, arXiv:2304.12244.

\bibitem[{Xu et~al.(2024)Xu, Jiang, Niu, Deng, Poovendran, Choi, and Lin}]{xu2024magpiealignmentdatasynthesis}
Zhangchen Xu, Fengqing Jiang, Luyao Niu, Yuntian Deng, Radha Poovendran, Yejin Choi, and Bill~Yuchen Lin. 2024.
\newblock \href {https://arxiv.org/abs/2406.08464} {Magpie: Alignment data synthesis from scratch by prompting aligned llms with nothing}.
\newblock \emph{Preprint}, arXiv:2406.08464.

\bibitem[{Yu et~al.(2023)Yu, Zhuang, Zhang, Meng, Ratner, Krishna, Shen, and Zhang}]{yu2023large}
Yue Yu, Yuchen Zhuang, Jieyu Zhang, Yu~Meng, Alexander~J Ratner, Ranjay Krishna, Jiaming Shen, and Chao Zhang. 2023.
\newblock Large language model as attributed training data generator: A tale of diversity and bias.
\newblock \emph{Advances in Neural Information Processing Systems}, 36:55734--55784.

\bibitem[{Zhou et~al.(2023)Zhou, Liu, Xu, Iyer, Sun, Mao, Ma, Efrat, Yu, YU, Zhang, Ghosh, Lewis, Zettlemoyer, and Levy}]{lima}
Chunting Zhou, Pengfei Liu, Puxin Xu, Srinivasan Iyer, Jiao Sun, Yuning Mao, Xuezhe Ma, Avia Efrat, Ping Yu, LILI YU, Susan Zhang, Gargi Ghosh, Mike Lewis, Luke Zettlemoyer, and Omer Levy. 2023.
\newblock \href {https://proceedings.neurips.cc/paper_files/paper/2023/file/ac662d74829e4407ce1d126477f4a03a-Paper-Conference.pdf} {Lima: Less is more for alignment}.
\newblock In \emph{Advances in Neural Information Processing Systems}, volume~36, pages 55006--55021. Curran Associates, Inc.

\end{thebibliography}

\onecolumn 
\newtcolorbox[auto counter, number within=section]{promptbox}[2][]{
    coltitle=white,
    fonttitle= \bfseries\ttfamily,
    sharp corners=south,
    title=Prompt~\thetcbcounter: #2,
    fontupper=\ttfamily,
    breakable
}

\appendix

\section{Examples}
\label{sec:appendix_examples}

This section presents an example of our data synthesis framework using \textsc{Reference-Level Feedback}. First, we select a reference sample:

\begin{promptbox}{Reference Sample}
\textbf{Instruction}: I have a string representing a unix timestamp (i.e.\ ``1284101485'') in Python, and I'd like to convert it to a readable date. When I use \texttt{time.strftime}, I get a \texttt{TypeError}:

>>>import time\\
>>>print time.strftime("\%B \%d \%Y", "1284101485")\\
Traceback (most recent call last):\\
  File "", line 1, in \\
TypeError: argument must be 9-item sequence, not str \\

\textbf{Response:}
There are two parts here:

\begin{itemize}
\item Convert the unix timestamp (``seconds since epoch'') to the local time
\item Display the local time in the desired format.
\end{itemize}

A portable way to get the local time that works even if the local time zone had a different utc offset in the past and python has no access to the tz database is to use a \texttt{pytz} timezone:\\

\#!/usr/bin/env python\\
from datetime import datetime\\
import tzlocal  \# \$ pip install tzlocal\\
unix\_timestamp = float("1284101485")\\
local\_timezone = tzlocal.get\_localzone() \# get pytz timezone\\
local\_time = datetime.fromtimestamp(unix\_timestamp, local\_timezone)\\

To display it, you could use any time format that is supported by your system e.g.:\\

print(local\_time.strftime("\%Y-\%m-\%d \%H:\%M:\%S.\%f\%z (\%Z)"))\\
print(local\_time.strftime("\%B \%d \%Y"))  \# print date in your format\\

If you do not need a local time, to get a readable UTC time instead:\\

utc\_time = datetime.utcfromtimestamp (unix\_timestamp)\\
print(utc\_time.strftime("\%Y-\%m-\%d \%H:\%M:\%S.\%f+00:00 (UTC)"))\\ 

If you don't care about the timezone issues that might affect what date is returned or if python has access to the tz database on your system:\\

local\_time = datetime.fromtimestamp (unix\_timestamp)\\
print(local\_time.strftime("\%Y-\%m-\%d \%H:\%M:\%S.\%f"))\\

On Python 3, you could get a timezone-aware datetime using only stdlib (the UTC offset may be wrong if python has no access to the tz database on your system e.g., on Windows):\\

\#!/usr/bin/env python3\\
from datetime import datetime, timezone\\
utc\_time = datetime.fromtimestamp (unix\_timestamp, timezone.utc)\\
local\_time = utc\_time.astimezone()\\
print(local\_time.strftime("\%Y-\%m-\%d \%H:\%M:\%S.\%f\%z (\%Z)"))\\

Functions from the \texttt{time} module are thin wrappers around the corresponding C API and therefore they may be less portable than the corresponding \texttt{datetime} methods otherwise you could use them too:\\

\#!/usr/bin/env python\\
import time\\
unix\_timestamp  = int("1284101485")\\
utc\_time = time.gmtime(unix\_timestamp)\\
local\_time = time.localtime(unix\_timestamp)\\
print(time.strftime("\%Y-\%m-\%d \%H:\%M:\%S", local\_time))\\
print(time.strftime("\%Y-\%m-\%d \%H:\%M:\%S+00:00 (UTC)", utc\_time))

\end{promptbox}

Then, we collect \textsc{Reference-Level Feedback} for that reference sample:
\begin{promptbox}{\textsc{Reference-Level Feedback}}
    \textbf{Instruction Subject Feedback:} Programming, specifically Python programming with a focus on date and time manipulation using Unix timestamps. \\

    \textbf{Instruction Skill Feedback:} Understanding of Unix timestamps, Python's datetime and time modules, exception handling in Python, formatting dates and times, knowledge of time zones and portability of code across different systems. \\

    \textbf{Response Feedback:} The reference response is effective in addressing the instruction for multiple reasons. Firstly, it accurately identifies the issue with the TypeError during the use of `time.strftime` and provides the correct method of converting a unix timestamp into a readable date format using the `datetime` module, which is more appropriate for this task. The response recognizes the importance of local time and considers timezone issues, which adds depth to the explanation.
    
    Additionally, the structure of the response is clear and organized into distinct sections that guide the user step-by-step through the conversion process. It provides multiple options (using different libraries and methods) for handling the conversion, catering to various user needs, which enhances comprehensiveness.
    
    However, there is room for improvement. 
    1. **Clarity**: While the response provides various methods, it could improve clarity by explicitly stating under what circumstances each method should be used (e.g., when to use `tzlocal`, when UTC is sufficient, etc.). 
    2. **Comprehensiveness**: The response could briefly explain what a Unix timestamp is for those unfamiliar with it and its relevance in this context. 
    3. **Engagement**: Incorporating a more conversational tone or additional commentary about best practices when dealing with date and time conversions could make the response feel more engaging. 
    4. **Potential errors**: It might be worth noting that `pytz` needs to be installed and that some users might run into issues if they don't set up their environment beforehand. 
    5. **Code snippets**: Ensure that code snippets are correctly formatted for clarity, especially in online platforms.
    
    Overall, the response effectively meets the instruction but could enhance user understanding and engagement with minor adjustments.
    
\end{promptbox}

Then, we begin the data synthesis. This process involves generating 10 new instructions for the instruction subject feedback and 10 for the instruction skill feedback. For each instruction, we generate a response and improve the response with the reference response feedback.

Below, we show an example of a new instruction that was generated based on the instruction skill feedback, as well as it's generated response and response improved with the reference response feedback.

\begin{promptbox}

\textbf{Instruction:} 
I am working with a Python script that reads timestamps in milliseconds from a file, but when I try to convert them to a readable format using \texttt{datetime.utcfromtimestamp}, I face a \texttt{TypeError}. My code looks like this:\\

import datetime\\
timestamps = [1640995200000, 1641081600000]\\
for ts in timestamps:\\
    print(datetime.utcfromtimestamp(ts)) \\

\textbf{Response:}
The issue you are encountering is due to the fact that \texttt{datetime.utcfromtimestamp} expects the timestamp to be in seconds, not milliseconds. Since your timestamps are in milliseconds, you need to convert them to seconds by dividing each timestamp by 1000 before passing it to \texttt{utcfromtimestamp}. Here's how you can modify your code to work correctly:\\

import datetime\\
timestamps = [1640995200000, 1641081600000]\\
for ts in timestamps:\\
    \# Convert milliseconds to seconds\\
    seconds = ts / 1000\\
    \# Get the UTC datetime from the seconds\\
    print(datetime. utcfromtimestamp(seconds))\\

In this code:
\begin{itemize}
\item We divide each timestamp by 1000 to convert from milliseconds to seconds.
\item Then, we call \texttt{datetime.utcfromtimestamp} with the converted value.
\end{itemize}

When you run this modified code, you should see the output in a readable format without encountering a \texttt{TypeError}:\\

2022-01-01 00:00:00\\
2022-01-02 00:00:00\\

This output represents the UTC datetime for January 1, 2022, and January 2, 2022, respectively. If you need to format the output differently, you can also use \texttt{strftime} method to customize the output format. For example:\\

for ts in timestamps:\\
    seconds = ts / 1000\\
    utc\_time = datetime.utcfromtimestamp(seconds)\\
    formatted\_time = utc\_time.strftime('\%Y-\%m-\%d \%H:\%M:\%S')\\
    print(formatted\_time) \\

\textbf{Improved Response: }
The issue you are facing with \texttt{TypeError} is due to the fact that the \texttt{datetime.utcfromtimestamp} function expects timestamps in seconds, not milliseconds. A Unix timestamp represents the number of seconds that have elapsed since January 1, 1970 (known as the Unix Epoch). In your case, the timestamps are in milliseconds, so we need to convert them to seconds by dividing each timestamp by 1000 before passing it to \texttt{utcfromtimestamp}. Here's how you can adjust your code to work correctly:\\

import datetime\\
timestamps = [1640995200000, 1641081600000]\\
for ts in timestamps:\\
    \# Convert milliseconds to seconds\\
    seconds = ts / 1000\\
    \# Get the UTC datetime from the seconds\\
    print(datetime. utcfromtimestamp(seconds))\\

In this modified code:
\begin{itemize}
\item Each timestamp is divided by 1000 to convert from milliseconds to seconds.
\item Then, we use \texttt{datetime.utcfromtimestamp} to get the UTC time.
\end{itemize}

When you run this code, you should see the output in a readable format without encountering any \texttt{TypeError}:\\

2022-01-01 00:00:00\\
2022-01-02 00:00:00\\

This output represents the UTC datetime for January 1, 2022, and January 2, 2022, respectively. If you require the date in a specific format, you can customize it using the \texttt{strftime} method. For instance:\\

for ts in timestamps:\\
    seconds = ts / 1000\\
    utc\_time = datetime.utcfromtimestamp(seconds)\\
    formatted\_time = utc\_time.strftime('\%Y-\%m-\%d \%H:\%M:\%S')\\
    print(formatted\_time)\\

Keep in mind that if you're working across different time zones and need to account for local time, you may want to explore libraries like \texttt{pytz}. Just remember that you will need to install it first and ensure your environment is set up correctly. Handling date and time conversions can often be tricky, so it's good practice to decide upfront whether you need UTC or local times based on the context of your application.

\end{promptbox}

After considering the relevant aspects from the reference response feedback, the improved response enhances understanding and engagement. Some changes include: adding an explanation of unix timestamps, explaining when to use UTC and when to use local time, adopting a more conversational tone, and including details of how to setup the environment.

\section{Prompt Templates}
\label{sec:appendix_prompt_templates}

This section presents the prompt templates we use throughout our pipeline.

\subsection{Prompts Used For Feedback Collection}

\begin{promptbox}{Reference-Level Instruction Feedback}
\# Task \\
Analyze the following instruction to extract key features that make it effective for instruction tuning. This analysis will help create high-quality instruction-response pairs for training language models to better follow instructions. \\

\# Context \\
The analysis will be used to create training data for instruction tuning language models. Focus on qualities that help language models learn to: \\
- Better understand user intentions \\
- Recognize instruction patterns \\
- Generate appropriate responses \\

\# Input \\
\{\{
  "instruction": \textcolor{blue}{\{instruction\}}, \\
  "reference\_response": \textcolor{blue}{\{reference\_response\}}, \\
\}\}

\# Output Format \\
\{\{
  "subject\_areas": str, \# This should be a description of the relevant subject areas and domains the instruction covers
  "relevant\_skills": str \# This should be a description of the relevant skills required to provide a good response to the instruction
\}\} \\

\# Analysis Guidelines \\
- Consider what makes this instruction clear and actionable \\
- Identify all relevant domains and skills \\
- Note structural elements that enhance instruction clarity \\

Output only a JSON object, in the format specified
\end{promptbox}

\begin{promptbox}{Response Feedback}
\# Task \\
Analyze the instruction-response pair and provide detailed feedback on how well it addresses the instruction. The feedback should: \\
- Highlight the specific qualities that make the response effective \\
- Provide actionable feedback for improvement \\

\# Input \\
\{\{
  "instruction": \textcolor{blue}{\{instruction\}}, \\
  "reference\_response": \textcolor{blue}{\{reference\_response\}}, \\
\}\} \\

\# Evaluation Criteria \\
\#\# Content Quality \\
- Accuracy and factual correctness \\
- Quality and depth of coverage \\

\#\# Communication \\
- Clarity and comprehensiveness \\
- Logical flow, organization, and structure \\
- Appropriate quality and depth \\
- Engagement and tone \\

\#\# Instruction Alignment \\
- How will it addresses the instruction \\
- Appropriate scope and focus \\
- Match with implied user needs \\

\# Output Format \\
\{\{
  "response\_feedback" : str \# Feedback describing strengths of the response and how it can be improved
\}\} \\

Output only a JSON object, in the format specified.\end{promptbox}

\subsection{Prompts Used For Data Synthesis}

\begin{promptbox}{Instruction Synthesis}
\# Task \\
Generate 10 new instructions based on the provided instruction feature and sample. Each instruction should: \\
- Be of similar complexity and length to the sample instruction \\
- Be practical and reasonable to answer \\
- Be diverse and high-quality \\

\# Sample Instruction: \\
\textcolor{blue}{\{instruction\}}

\# Instruction Features: \\
\textcolor{blue}{\{feature\}} \\

\# Output Format \\
\{\{
  "instructions": list \# List of 10 distinct instructions. Each instruction should be a single string.
\}\} \\

Output only a JSON object, in the format specified.
\end{promptbox}

\begin{promptbox}{Response Synthesis}
\# Task \\
I will provide an instruction. Generate a high-quality, helpful response to the instruction. The response should demonstrate expertise, clear reasoning, and natural language use. \\

\# Response Requirements \\
- Directly address all aspects of the instruction \\
- Response should demonstrate clear reasoning and expertise \\
- Use clear, natural language \\
- Include examples or evidence when relevant \\
- Show step-by-step reasoning where appropriate \\
- Maintain appropriate length and detail level \\
- Use proper formatting (lists, paragraphs) as needed \\

Here is an example of a response to an instruction: \\
\# Sample Input Instruction:
\textcolor{blue}{\{sample\_instruction\}} \\
\# Sample Response: \\
\textcolor{blue}{\{reference\_response\}} \\

\# Output Format \\
\{\{
  "response": "The complete response text here"
\}\}

\# Input \\
\{\{
  "instruction": \textcolor{blue}{\{instruction\}},
\}\} \\

Generate a properly formatted JSON response, as specified by the Output Format, that addresses this instruction.
\end{promptbox}

\begin{promptbox}{Response Refinement with Reference-Level Feedback}
\# Task \\
Given an instruction-response pair and feedback, generate an improved version of the response by applying the feedback. The feedback was given for a similar but different instruction-response pair. Not all aspects of the feedback may be directly applicable, so make sure to only apply relevant aspects of the feedback. \\

\# Input \\
\{\{
  "instruction": \textcolor{blue}{\{instruction\}},
  "original\_response": \textcolor{blue}{\{response\}},
  "feedback": \textcolor{blue}{\{response\_feedback\}}
\}\} \\

\# Quality Assessment Process \\
1. Analyze Original Response \\
- Core strengths and effective elements \\
- Structure and organization \\
- Depth and comprehensiveness \\
- Alignment with instruction \\

2. Evaluate Feedback \\
- Identify feedback points that are relevant to improving this response, and ignore points that are not relevant \\
- Identify actionable improvement suggestions \\
- Assess potential impact of each change \\
- Check alignment with original instruction \\
- Validate that suggested changes maintain or enhance quality \\

3. Improvement Strategy \\
- Prioritize changes with highest impact \\
- Preserve effective elements of the original response \\
- Ensure feedback applied enhance the response and do not remove valuable elements \\

\# Output Format  \\
\{\{
    "analysis": \{\{ \\
        "original\_strengths": ["list of key effective elements to preserve"],
        "improvement\_opportunities": ["list of specific areas that will benefit from enhancement"],
        "relevant\_feedback": ["list of feedback points that are relevant and beneficial"]
    \}\},
    "implementation\_strategy": \{\{ \\
        "planned\_changes": ["identify what feedback will be applied"],
        "rationale": "explain how this feedback will improve the original response"
    \}\},
    "improved\_response": "The revised and improved response"
\}\} \\

Output only a JSON object, in the format specified.
\end{promptbox}

\begin{promptbox}{Response Refinement with Sample-Level Feedback}
\# Task \\
Given an instruction-response pair and feedback, generate an improved version of the response by applying the feedback. \\

\# Input \\
\{\{
  "instruction": \textcolor{blue}{\{instruction\}},
  "original\_response": \textcolor{blue}{\{response\}},
  "feedback": \textcolor{blue}{\{self\_reflection\}}
\}\} \\

\# Quality Assessment Process \\
1. Analyze Original Response \\
- Core strengths and effective elements \\
- Structure and organization \\
- Depth and comprehensiveness \\
- Alignment with instruction \\

2. Evaluate Feedback \\
- Identify actionable improvement suggestions \\
- Assess potential impact of each change \\
- Check alignment with original instruction \\
- Validate that suggested changes maintain or enhance quality \\

3. Improvement Strategy \\
- Prioritize changes with highest impact \\
- Preserve effective elements of the original response \\
- Ensure feedback applied enhance the response and do not remove valuable elements \\

\# Output Format \\
\{\{
    "analysis": \{\{ \\
        "original\_strengths": ["list of key effective elements to preserve"],
        "improvement\_opportunities": ["list of specific areas that will benefit from enhancement"]
    \}\},
    "implementation\_strategy": \{\{ \\
        "planned\_changes": ["identify what feedback will be applied"],
        "rationale": "explain how this feedback will improve the original response"
    \}\},
    "improved\_response": "The revised and improved response"
\}\} \\

Output only a JSON object, in the format specified.
\end{promptbox}

\section{Scalability}
\label{sec:scalability}

\begin{table}[H]
    \centering
    \small
    \setlength{\tabcolsep}{4pt}
    
    \begin{tabular}{c l c c c c c c c} \toprule
        \multicolumn{2}{c}{\multirow{2}[3]{*}{\makecell{\textbf{Model} }}}
        & {\multirow{2}[3]{*}{{\textbf{\# Samples}}}}
        & \multicolumn{4}{c}{\textbf{AlpacaEval 2}}
        & \multicolumn{2}{c}{\textbf{Arena-Hard}} 
        \\ 
        \cmidrule(lr){4-7} \cmidrule(lr){8-9}
        & & & LC (\%) & WR (\%) & SE & Len. & WR (\%) & Tok. \\ \midrule        
        
        & \textbf{Llama-3.1-8B-Instruct-\textsc{REFED}} & 10K & 43.96  &  42.24  & 1.74  & 1950  & 35.9 & 670\\
        & \textbf{Llama-3.1-8B-Instruct-\textsc{REFED-20K}} & \textbf{20K} & \textbf{45.53}  &  \textbf{44.60}  &  \textbf{1.75}  &  \textbf{805}  &  \textbf{42.3}  &  \textbf{705}\\
        \bottomrule
    \end{tabular}
    
    \caption{ Evaluation results of Llama-3.1-8B-Instruct finetuned on \textsc{REFED} and \textsc{REFED-20K}. Results demonstrate promising results when scaling data synthesis with \textsc{Reference-Level Feedback}, with significant improvements as dataset size increases.
    }
    \label{tab:table5}
\end{table}

\section{Human Evaluation}
\label{sec:human_eval}

\begin{promptbox}{Instructions Provided to Human Annotators}

\textbf{Instructions for Response Preference Annotation}

Our human evaluation study was conducted with expert annotators recruited from university programs. We randomly select 50 instructions, and two corresponding responses. We ask them to select their preferred response. The prompt we provide to them is below.

Thank you for participating in this human annotation! \\

You will be reviewing pairs of AI-generated responses to an instruction. For each instruction, two different responses are provided. Your task is to indicate which response you prefer.
Please scroll through this document, and indicate your preferred response for each instruction by writing either “1” or “2” in the blank space provided. \\

\textbf{Notes and Guidelines}
\begin{itemize}
    \item Trust your judgement! This is a subjective task and there are no right/wrong answers
    \item Consider factors such as accuracy, helpfulness, clarity, structure, and engagement…
\end{itemize}

\end{promptbox}

\section{License}
Our use of existing artifact(s) is consistent with their intended use. The LIMA dataset follows the CC BY-NC-SA license, or a stricter license if the source data follows the same.

\end{document}